%% file: main.tex
\documentclass{article}

\usepackage{arxiv}
\usepackage{placeins}
\usepackage[utf8]{inputenc} % allow utf-8 input
\usepackage[T1]{fontenc}    % use 8-bit T1 fonts
\usepackage[scaled=1.0,varqu,varl]{zi4} % Inconsolata as \ttfamily: strong, clearly
\usepackage{hyperref}       % hyperlinks
\usepackage{url}            % simple URL typesetting
\usepackage{booktabs}       % professional-quality tables
\usepackage{amsfonts}       % blackboard math symbols
\usepackage{nicefrac}       % compact symbols for 1/2, etc.
\usepackage{microtype}      % microtypography
\usepackage{lipsum}
\usepackage{graphicx}
\graphicspath{ {./images/} }
\usepackage{subcaption}
\usepackage{booktabs}
\usepackage{mathtools}
\usepackage{wrapfig}
\usepackage{hyperref}
\usepackage{makecell}
\usepackage{xcolor}
\usepackage{tabularx}
\usepackage[nolist]{acronym}
\usepackage[style=numeric, maxnames=99, sorting=none, backend=bibtex]{biblatex}
\AtEveryBibitem{
    \clearfield{note}
    \clearfield{extra}
    \clearfield{url}
    \clearfield{extra}
    \clearfield{urlyear}
    \clearfield{urlmonth}
    \clearfield{urlday}
    \clearlist{urldate}}
\AddToHook{begindocument/end}{\pdfstringdefDisableCommands{%
  \def\ac#1{#1}\def\acs#1{#1}\def\acl#1{#1}\def\acf#1{#1}%
  \def\acp#1{#1s}}}

\usepackage[subdued]{mathastext}
\MTencoding{T1}\MTfamily{zi4}\MTseries{m}
\Mathastext[algott]

\newcounter{algorithm}

\newenvironment{algorithm}[1][htbp]
  {\refstepcounter{algorithm}\begin{table}[#1]\renewcommand{\tablename}{Algorithm}%
   \renewcommand{\thetable}{\thealgorithm}}
  {\end{table}\addtocounter{table}{-1}}% the caption must not advance the table counter

\title{A Human-in-the-Loop Bayesian Optimization Framework for Constraint-Aware Bioprocess Development}

\author{
  Samuel Stricker \\
  Sargent Centre, Department of Chemical Engineering\\
  Imperial College London, London, UK \\
  \And
  Claus Wirnsperger \\
  DataHow AG\\
  Zurich, Switzerland \\
  \And
  Alessandro Butt\'e\\
  DataHow AG\\
  Zurich, Switzerland \\
  \And
  Laura Marie Helleckes \\
  Sargent Centre, Department of Chemical Engineering\\
  and I-X Centre for AI in Science\\
  Imperial College London, London, UK \\
  \And
  Gonzalo Guill\'en Gos\'albez \\
  Department of Chemistry and Applied Biosciences\\
  ETH Zurich, Zurich, Switzerland \\
  \And
  Antonio del Rio Chanona \\
  Sargent Centre, Department of Chemical Engineering\\
  Imperial College London, London, UK \\
  \And
  Mehmet Mercang\"oz\thanks{Corresponding author: \texttt{m.mercangoz@imperial.ac.uk}} \\
  Sargent Centre, Department of Chemical Engineering\\
  Imperial College London, London, SW7 2AZ, UK \\
}

\begin{document}

\begin{acronym}
    \acro{PFGS}{Pareto Front Guided Sampling}
    \acro{HitL}[HiL]{Human-in-the-Loop}
    \acro{BO}{Bayesian Optimization}
    \acro{GP}{Gaussian process}
    \acroplural{GP}[GPs]{Gaussian processes}
    \acro{CHO}{Chinese Hamster Ovary}
    \acro{CQA}{critical quality attribute}
    \acro{CPP}{critical process parameter}
    \acro{QbD}{Quality by Design}
    \acro{ICH}{International Council for Harmonization}
    \acro{DoE}{Design of Experiments}
    \acro{ARBO}{adversarially robust BO}
    \acro{EI}{Expected Improvement}
    \acro{PI}{Probability of Improvement}
    \acro{CDF}{cumulative distribution function}
    \acro{PDF}{Probability Density Function}
    \acro{UCB}{Upper Confidence Bound}
    \acro{MC}{Monte Carlo}
    \acro{NSGAII}[NSGA-II]{Non-dominated Sorting Genetic Algorithm II}
    \acro{LHS}{Latin hypercube sampling}
    \acro{MOEEQI}[MO-E-EQI]{multi-objective Euclidean expected quantile improvement}
\end{acronym}

\maketitle
\begin{abstract}
We present \acf{PFGS}, a human-in-the-loop (HiL) \acf{BO} framework that reformulates surrogate-model predictions as a multi-objective optimization problem, allowing domain knowledge to be added intuitively to any \ac{BO} problem. \ac{PFGS} addresses constrained optimization by treating the posterior probability of constraint satisfaction as an explicit Pareto objective, and robust optimization via a Monte Carlo estimate of expected lower-confidence performance under input perturbations, capturing likely degradation. The multi-dimensional Pareto front makes trade-offs between predicted performance, model uncertainty, constraint satisfaction, and robustness visible through pairwise two-dimensional projections on an interactive dashboard, allowing selection criteria to be refined as the model and objectives evolve. The framework is showcased on an eight-dimensional in-silico fed-batch Chinese Hamster Ovary cell culture case study. We showed improved performance over \ac{BO} when the expert has perfect domain knowledge; however, \ac{PFGS} is not designed to outperform \ac{BO}, but rather to present a HiL acquisition function framework.
\end{abstract}

\acresetall % the abstract is self-contained: expand acronyms again at first use in the main text
\newpage
\input{01_introduction}

\input{02_methods}
\input{02.1_CaseStudies}
\input{03_results}
\input{04_conclusion}
\input{05_acknowledgments}

\printbibliography

\clearpage
\input{06_appendix}

\end{document}

%% file: 01_introduction.tex
\section{Introduction}
    Chemical and process engineering problems are commonly characterized by high-dimensional operating spaces, expensive experimentation, nonlinear process interactions, and stringent operational constraints. Process development therefore requires methodologies that can efficiently extract maximal information from limited experimental campaigns while simultaneously accounting for uncertainty, feasibility, and robustness~\cite{siska_guide_2026}. These requirements arise across a broad range of engineering applications, including catalysis~\cite{shields_bayesian_2021}, materials synthesis~\cite{frazier_bayesian_2016}, separations~\cite{perez-ones_stochastic_2024}, and biomanufacturing~\cite{siska_guide_2026}.
    
    Historically, \ac{DoE} has provided the statistical foundation for systematic process characterization and optimization. Classical experimental designs, like factorial and fractional factorial schemes, as well as response-surface methodologies such as central composite designs, enable efficient estimation of main effects, interactions, and higher-order curvature terms from relatively small experimental campaigns~\cite{kasemiire_design_2021}. Consequently, \ac{DoE} methodologies remain central tools for factor screening, sensitivity analysis, and process understanding throughout chemical and pharmaceutical development. However, conventional \ac{DoE} strategies are fundamentally static: the experimental matrix is defined \emph{a priori}, preventing adaptation as information accumulates during experimentation. This limitation becomes increasingly restrictive in complex systems where experiments are costly, noisy, or poorly understood initially~\cite{greenhill_bayesian_2020}. %Additionally, classical methods often use assumptions that do not hold in \ac{PSE}.

    To address these challenges, sequential and data-efficient optimization frameworks have gained increasing attention. Among these, \ac{BO} has emerged as a particularly powerful methodology for experimental design and process optimization in expensive black-box systems~\cite{frazier_tutorial_2018}. \ac{BO} constructs a probabilistic surrogate model of the response surface and sequentially proposes new experiments by optimizing an acquisition function that balances exploration of uncertain regions with exploitation of promising operating conditions. This uncertainty-aware paradigm is especially attractive in engineering settings where each experiment or simulation carries substantial economic cost and is often extremely time-consuming~\cite{frazier_bayesian_2016}.

    This work investigates the concepts of \ac{BO} in the context of bioprocess engineering, a domain that strongly amplifies the general challenges of process optimization due to high experimental cost, intrinsic biological variability, and stringent regulatory requirements. However, in this field there is diverse expert knowledge that could guide experimentation more efficiently. Therefore, \ac{HitL} frameworks are promising for human-algorithm collaboration. In \ac{HitL} methodologies the goal is to leverage the complementary strengths of human intuition and algorithmic efficiency to improve decision-making in complex optimization tasks. By integrating domain expertise into the optimization loop, \ac{HitL} approaches can enhance the interpretability, trustworthiness, and overall performance of \ac{BO} frameworks, especially in high-stakes or highly regulated environments where expert judgment is essential for assessing feasibility, safety, and regulatory compliance~\cite{savage_human-algorithm_2024}.
    
    Within biopharmaceutical manufacturing, \ac{QbD} has become the prevailing scientific and regulatory framework for process development, wherein product quality is prospectively engineered rather than retrospectively tested~\cite{yang_aspects_2025}, which closely aligns with the core advantages of \ac{BO}. In \ac{QbD} \acp{CQA} and \acp{CPP} are systematically identified and connected according to the \ac{ICH} Q8--Q11 guidelines~\cite{yang_aspects_2025}. These relationships define the process design space and associated control strategy that underpin robust and flexible manufacturing operations. Given the high process complexity and stringent regulatory requirements, the \ac{HitL} paradigm is essential, as domain experts can directly encode process requirements into the optimization loop. This work introduces the \ac{PFGS} algorithm which can incorporate \ac{CQA} constraints and \ac{CPP} variability into \ac{BO}, enabling robust and constrained optimization.

    \subsection{Background \& Related Works}
        There are different ways to design and optimize processes, in this section we describe the main ideas and highlight related research within the fields of biotechnology and chemical engineering.

        \subsubsection{Classical Design of Experiments}
            Classical \ac{DoE} provides a structured framework for selecting experimental conditions so that an empirical model of a process can be built from a limited number of observations. As summarized by Greenhill et al., traditional \ac{DoE} methods seek to measure the design space efficiently, thereby supporting prediction, process understanding, and subsequent optimization when experiments are expensive~\cite{greenhill_bayesian_2020}. In this setting, the experimental plan is usually specified before the campaign begins, and the choice of sampling strategy is closely linked to the type of model that is expected to describe the response.

\begin{figure}[h]
            \centering
            \vspace{-0.6cm}
            \begin{subfigure}[b]{0.31\textwidth}
                \centering
                \includegraphics[width=0.7\textwidth]{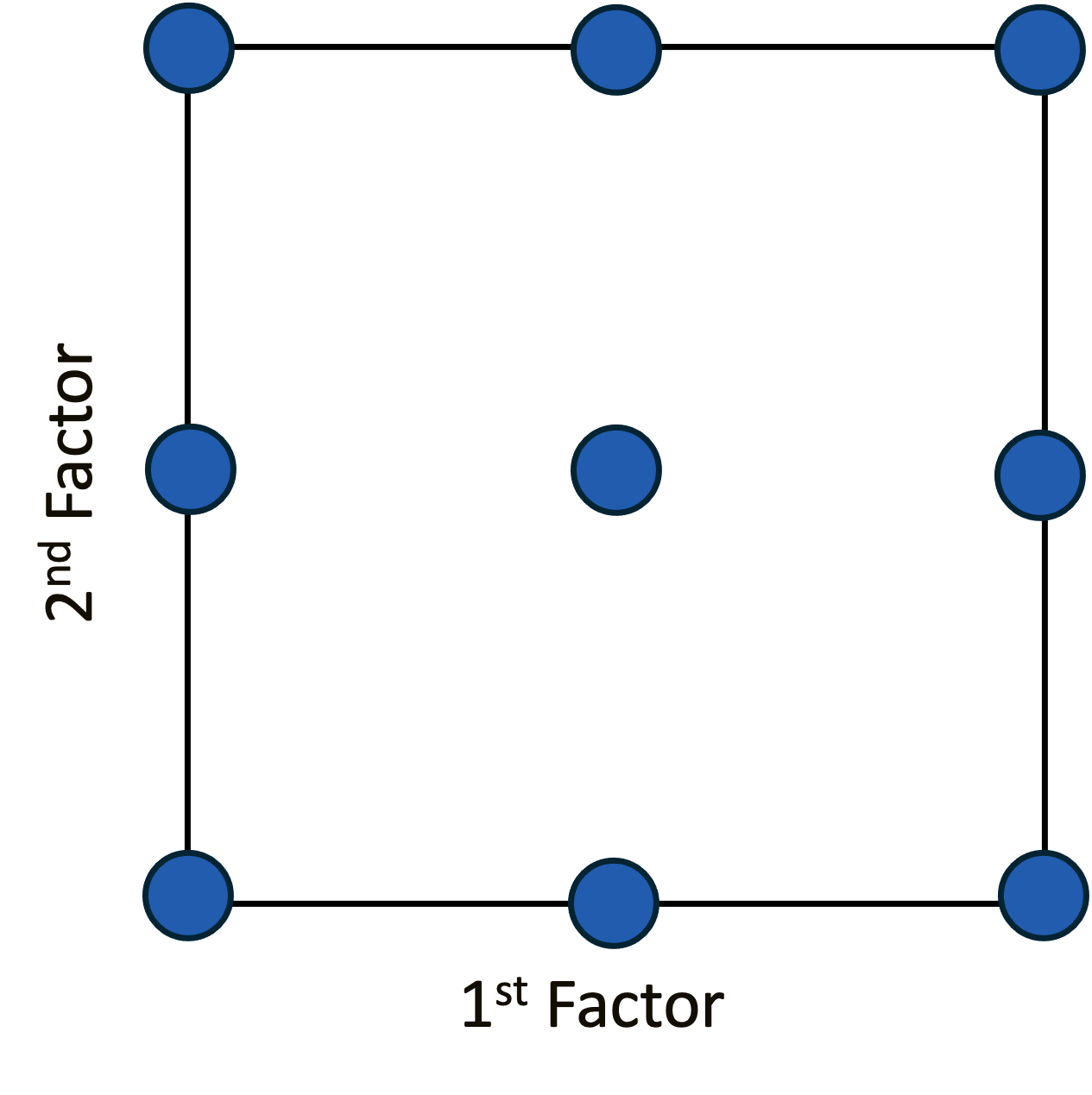}
                \caption{Factorial design}
                \label{fig:doe_factorial}
            \end{subfigure}
            \hfill
            \begin{subfigure}[b]{0.31\textwidth}
                \centering
                \includegraphics[width=\textwidth,trim=30pt 0 30pt 0,clip]{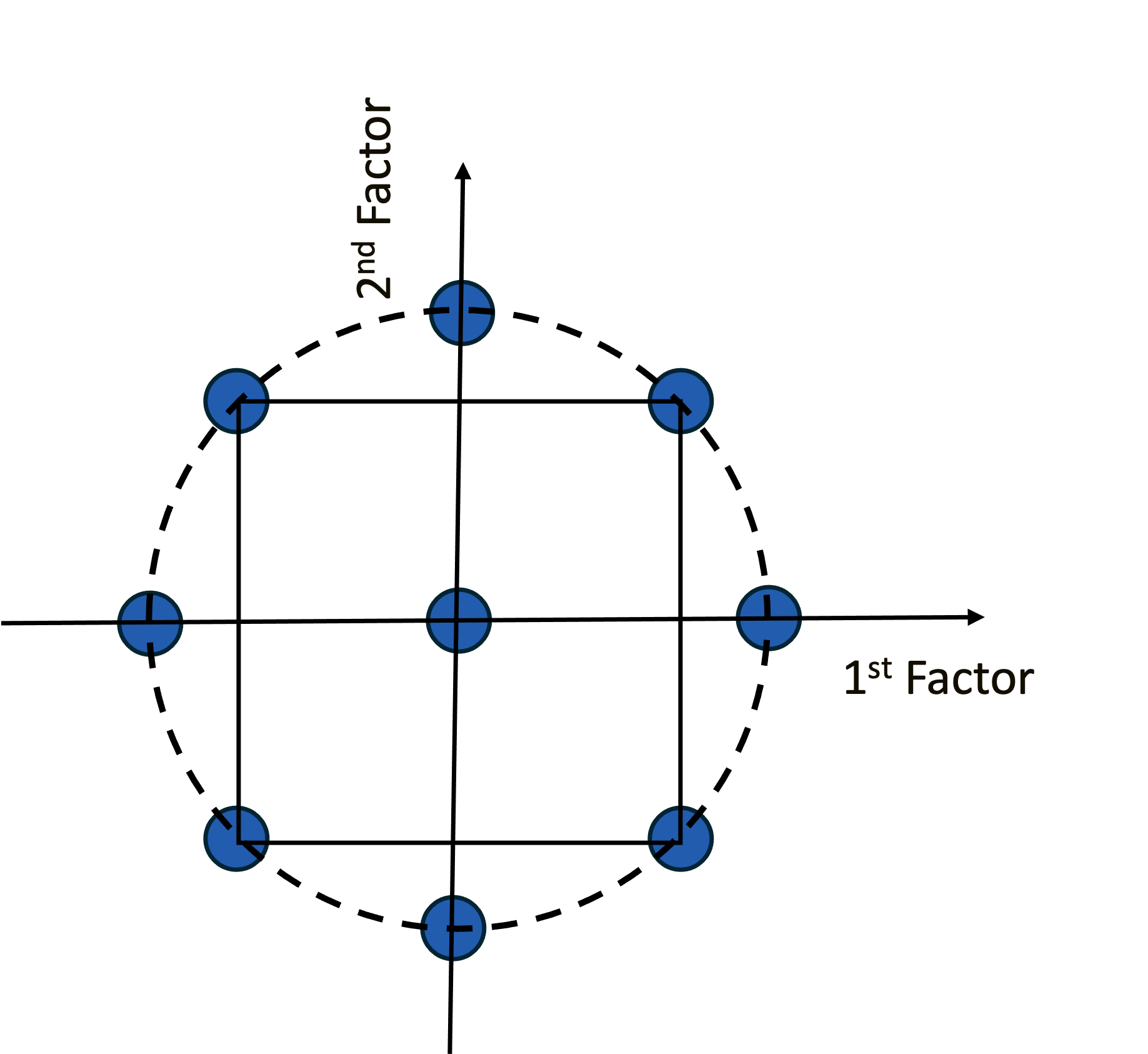}
                \caption{Central composite design}
                \label{fig:doe_ccd}
            \end{subfigure}
            \hfill
            \begin{subfigure}[b]{0.31\textwidth}
                \centering
                \includegraphics[width=0.7\textwidth]{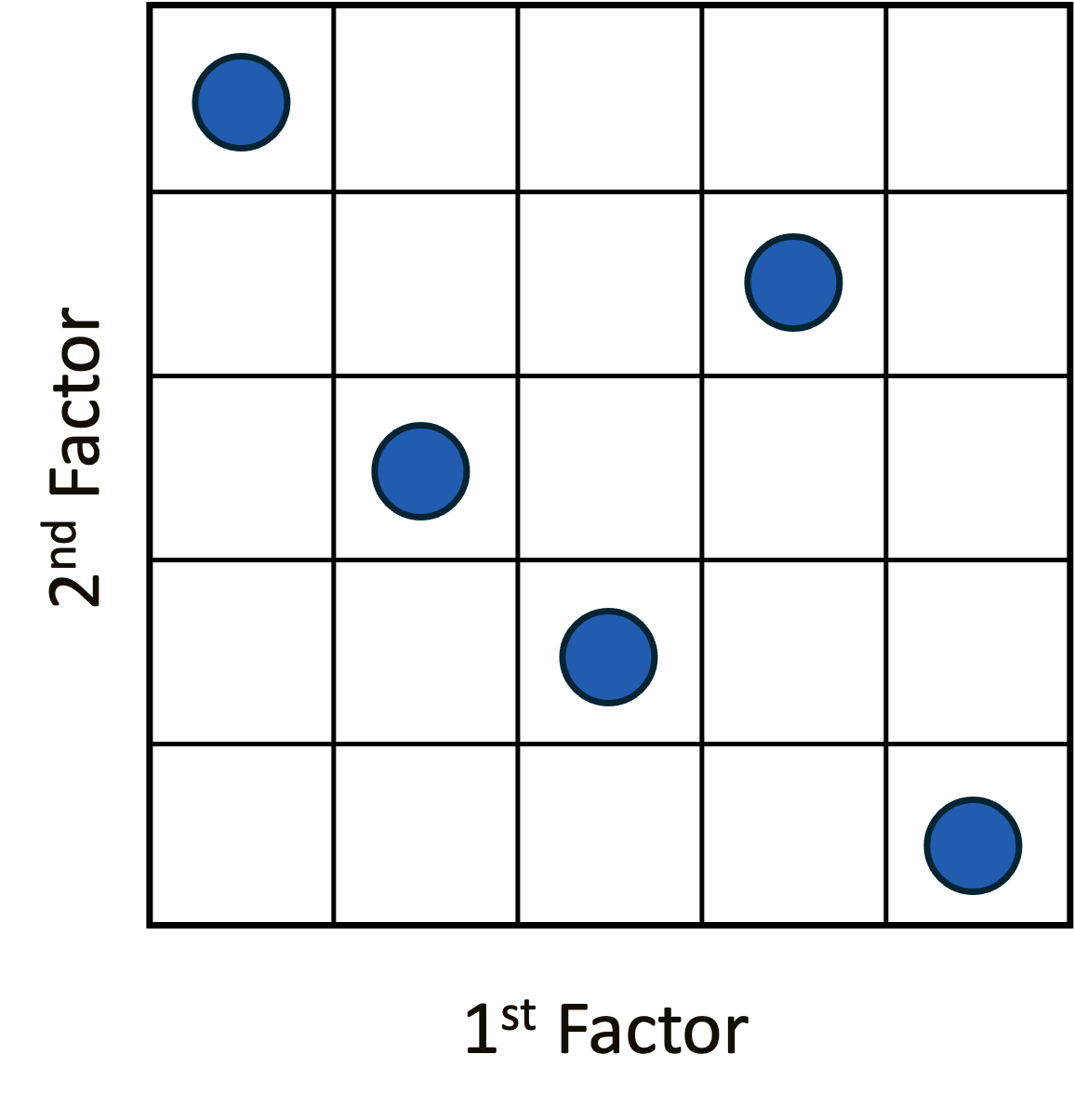}
                \caption{Latin hypercube design}
                \label{fig:doe_lhs}
            \end{subfigure}
            \caption{Representative classical experimental design strategies. (a) Factorial designs sample combinations of factor levels on a structured grid, enabling estimation of main effects and interactions. (b) Central composite designs augment factorial points with axial and center points to estimate response-surface curvature. (c) Latin hypercube designs distribute samples across the design space to improve space filling when less is known about the response surface.}
            \label{fig:classical_doe_designs}
        \end{figure}

            Factorial designs shown in Figure~\ref{fig:doe_factorial} are among the most established classical \ac{DoE} approaches. They place samples on a structured grid and are well suited to estimating main effects and interactions when a low-order model is appropriate. Full factorial designs evaluate all combinations of selected factor levels, which gives rich information about interactions but becomes costly as the number of variables increases. Fractional factorial and screening designs reduce this burden by sampling only part of the full grid, making them useful for identifying influential variables before more detailed optimization is attempted~\cite{greenhill_bayesian_2020,kasemiire_design_2021}.
            
            For nonlinear systems, classical \ac{DoE} is often extended using response-surface and space-filling designs. Response-surface methodologies, such as central composite (Figure~\ref{fig:doe_ccd}) or Box--Behnken designs, fit polynomial approximations that can describe curvature and support local optimization of selected responses. Space-filling designs, including \ac{LHS} (Figure~\ref{fig:doe_lhs}), aim to cover the design space more uniformly and are useful when less is known about the response surface in advance. These approaches are more flexible than simple grids, but the required number of samples still increases rapidly with dimensionality, making high-dimensional experimental design challenging~\cite{greenhill_bayesian_2020}.

            A key limitation of classical \ac{DoE}, however, is that the experimental matrix is generally fixed \emph{a priori}, as modeling and optimization are usually treated as separate stages: first a model is built from a predefined set of experiments, and only then is the model used for optimization~\cite{greenhill_bayesian_2020}.

        \subsubsection{\texorpdfstring{\acl{BO}}{Bayesian Optimization}}
            \ac{BO} is a sequential, derivative free strategy for the global optimization of expensive black-box functions lacking analytical expressions. \ac{BO} uses probabilistic surrogate models like \acp{GP} trained  on the accumulated data and employs an acquisition function to select the next experimental condition \cite{frazier_tutorial_2018}. This paradigm is well-suited to experimental sciences where evaluations are costly. The \ac{BO} loop comprises three steps: (i)~fit a surrogate to observations $\mathcal{D}_{t} = \{(\mathbf{x}_{i}, y_{i})\}_{i=1}^{n_d}$ (Section~\ref{sec:gp}); (ii)~optimize an acquisition function $\alpha(\mathbf{x} \mid \mathcal{D}_{t})$ to propose the next point (Section~\ref{sec:aquisition}); (iii)~perform the proposed experiment and augment the dataset. This procedure is iterated until convergence or a given experimental budget is exhausted. A more detailed explanation of \ac{BO} and its variants can be found in~\ref{appx:BO}.
            
            Early demonstrations in bioprocess engineering focused on reducing experimental burden in cell culture: Mehrian et al.~\cite{mehrian_maximizing_2018} optimized cell-growth conditions in bioreactor culture, while Narayanan et al.~\cite{narayanan_design_2021} applied batch and multi-objective \ac{BO} to biopharmaceutical formulation development, identifying optimal antibody formulations with as few as 25 experiments. Subsequent work expanded the scope to upstream media design~\cite{cosenza_multi-objective_2023} and high-throughput protein engineering, where Helleckes et al.~\cite{helleckes_high-throughput_2024} coupled laboratory automation with \ac{BO} to navigate large combinatorial sequence spaces. Most recently, Martens et al.~\cite{martens_holistic_2025} extended the paradigm to multi-scale process development via a multi-fidelity batch \ac{BO} framework that jointly optimizes reaction conditions and biocatalyst selection across scales from microtiter plates to pilot reactors.

        \subsubsection{Constrained \& Safe \texorpdfstring{\acl{BO}}{Bayesian Optimization}}
            Many optimization problems are defined not only by a performance objective, but also by feasibility requirements. Two related paradigms address this setting, differing mainly in \emph{when} feasibility or safety must hold. Constrained \ac{BO} aims to identify a final solution that satisfies the constraints, whereas safe \ac{BO} requires every evaluated query during the optimization campaign to satisfy the safety constraints with high probability. A basic version of constrained \ac{BO} is explained in detail in~\ref{sec:constrained_bo}.
            
            Gardner et al.~\cite{gardner_bayesian_2014} formalized constrained \ac{BO} by placing independent Gaussian-process priors on both the objective and the constraint functions. In chemical and process engineering, Hickman et al.~\cite{hickman_bayesian_2022} demonstrated constrained \ac{BO} for the optimization of the synthesis of o-xylenyl adducts of Buckminsterfullerene, handling known experimental and design constraints within the Gryffin framework. Paulson and Lu~\cite{paulson_cobalt_2022} extended the setting to gray-box process models with expensive evaluations in their COBALT algorithm, which exploits derivative information to satisfy explicit inequality constraints efficiently. González and Zavala~\cite{gonzalez_implementation_2025} address the related composite-function setting, in which the objective is a known function of \ac{GP}-modeled intermediates; their BOIS framework uses adaptive linearizations to obtain analytical moments of the composite, avoiding the sampling or augmented-space formulations used by earlier gray-box methods at substantially lower computational cost. Kudva et al.~\cite{kudva_constrained_2022} further developed a constrained robust \ac{BO} framework with guaranteed sublinear regret bounds applicable to noisy black-box process design problems. Ahmed et al.~\cite{ahmed_arrtoc_2025} propose ARRTOC, an adversarially robust real-time optimization (RTO) algorithm that finds RTO set-points explicitly robust to control-layer implementation errors (disturbances/noise) by solving a constrained adversarial robust optimization problem via neighborhood exploration and SOCP-based robust local moves, demonstrating on bioreactor, evaporator, and plant-model-mismatch case studies that ARRTOC-selected set-points improve realized RTO profit by up to 50\% and substantially reduce constraint violations compared to nominal RTO.
    
        % \subsubsection{Safe \texorpdfstring{\acl{BO}}{Bayesian Optimization}}
            Safe \ac{BO} tightens the constraint-handling requirement by demanding that every queried point satisfy safety conditions with high probability, even during exploration~\cite{berkenkamp_bayesian_2023}. Krishnamoorthy and Doyle~\cite{krishnamoorthy_safe_2022} proposed model-free real-time optimization of process systems in which both the objective and constraints are unknown and must be learned from plant measurements, guaranteeing constraint satisfaction throughout the optimization trajectory. Petsagkourakis et al.~\cite{petsagkourakis_safe_2021} addressed safe real-time optimization under plant--model mismatch by combining multi-fidelity \acp{GP} with chance-constraint satisfaction, ensuring feasibility despite structural model uncertainty. Garces et al.~\cite{garces_efficient_2024} reported an approximately 80\% reduction in constraint violations when applying safe \ac{BO} to isotope-separation tuning. Petsagkourakis et al.~\cite{petsagkourakis_safe_2021} propose a trust-region-constrained Bayesian real-time optimization framework using multi-fidelity \acp{GP} with Chebyshev-Cantelli chance constraints, demonstrating on a photobioreactor batch-to-batch optimization case study that this approach converges faster and violates fewer constraints than expectation-based, no-trust-region, or model-free alternatives.

        \subsubsection{Robust \texorpdfstring{\acl{BO}}{Bayesian Optimization}}
            Classical \ac{BO} seeks an input with high predicted objective value, but the resulting optimum may be sensitive to small deviations between the proposed and realized operating conditions. This is problematic in experimental process development, where implementation errors, raw-material variability, or scale-dependent effects can perturb the selected recipe. Robust \ac{BO} therefore modifies the optimization goal from finding a single high-performing point to finding a point whose performance remains high under a specified perturbation set as shown more detailed in~\ref{sec:rob_bo}.
        
            Robust \ac{BO} addresses performance degradation under uncertainty that cannot be eliminated by constraint satisfaction alone. Le and Branke~\cite{phuong_le_bayesian_2020} define a taxonomy of robustness in terms of expected performance under disturbances and output noise, while Bogunovic et al.~\cite{bogunovic_adversarially_2018} and Weichert et al.~\cite{weichert_robust_2024} consider adversarial and entropy-search variants for perturbed inputs. Paulson et al.~\cite{paulson_adversarially_2022} introduced \ac{ARBO} for auto-tuning of control structures in chemical processes under time-invariant parametric uncertainty, demonstrating that a minimax formulation provides substantially stronger robustness guarantees than standard expected-improvement approaches. Kudva et al.~\cite{kudva_robust_2024} developed BoFlex, a robust \ac{BO} framework for worst-case flexibility analysis of simulation-based process models, providing rigorous uncertainty bounds and finite-iteration convergence guarantees. Kudva and Paulson~\cite{kudva_bonsai_2025} further proposed BONSAI, which exploits the network structure of modular process flowsheets to scale robust \ac{BO} to high-dimensional interconnected systems. Zhang et al.~\cite{zhang_multi-objective_2025} addressed multi-objective reaction optimization under heteroscedastic experimental noise via a \ac{MOEEQI} acquisition function, applied to an esterification case with competing yield and E-factor objectives.

        \subsubsection{Human-in-the-Loop \texorpdfstring{\acl{BO}}{Bayesian Optimization}}
            Despite the increasing autonomy of derivative free optimization frameworks, engineering process development remains fundamentally knowledge-driven. Expert intuition is often essential for incorporating considerations that are difficult to encode explicitly, including manufacturability, scale-up feasibility, safety, or regulatory concerns~\cite{romero-obon_human---loop_2025}. \ac{HitL} \ac{BO} has therefore emerged as an active research direction, particularly in low-data regimes where domain knowledge can substantially boost performance. Savage and del Rio Chanona et al.~\cite{savage_human-algorithm_2024} proposed expert-guided batch \ac{BO} using Pareto-based candidate selection, enabling domain experts to influence critical early decisions. Adachi et al.~\cite{adachi_looping_2024} introduced CoExBO, a collaborative and explainable \ac{BO} framework with a no-harm guarantee, validated on lithium-ion battery design. Kanarik et al.~\cite{kanarik_humanmachine_2023} empirically demonstrated that expert knowledge enhances \ac{BO} convergence during the initial exploration phase, though purely algorithmic \ac{BO} proved more efficient in later stages.

    \subsection{Research Gap \& Contributions}
        \ac{BO} has been applied to a wide range of problems in chemical and bioprocess engineering, yet its practical adoption remains limited by insufficient integration of human expertise and the consequent lack of trust in automated recommendations. \ac{HitL} frameworks are therefore particularly important in high-risk or highly regulated sectors such as biopharmaceutical manufacturing, where expert judgment is essential for assessing feasibility, safety, and regulatory relevance. The objective of this work is not to develop a novel acquisition function, but rather to reorganize existing acquisition strategies within a \ac{HitL} multi-objective optimization framework, making model-generated recommendations more interpretable and actionable for domain experts. This is especially relevant for navigating high-dimensional, noisy, and constrained bioprocess design spaces where purely automated optimization may overlook practical considerations that are difficult to encode explicitly.

        This work presents the \ac{HitL} \ac{BO} workflow of \ac{PFGS} preliminary developed in~\cite{stricker_optimal_2024,stricker_pareto_2026}. A introduction into \ac{PFGS} can be found in Section \ref{sec:pfgs}. The  main contributions are:
        \begin{itemize}
          \item \ac{PFGS} for constrained optimization by incorporating probabilistic \ac{CQA} satisfaction as an additional Pareto objective.
          \item Integration of robustness criteria that allow candidate recipes to be evaluated with respect to resilience under expected \ac{CPP} variability.
          \item Development of an interactive dashboard and visualization workflow that renders model recommendations interpretable and actionable for domain experts.
          \item Demonstration that existing \ac{BO} objectives can be reorganized into a \ac{HitL} multi-objective framework without requiring new acquisition function design.
        \end{itemize}
        Although the focus is on bioprocess engineering, the underlying methodologies are broadly transferable across engineering disciplines involving expensive experimentation, constrained optimization, and uncertainty-aware decision making.

%% file: 02_methods.tex
\section{Methods}
    This section describes the methodological framework developed to guide experimental recipe selection with \ac{GP} surrogate models and \ac{PFGS}. First, the basic \ac{PFGS} workflow is introduced based on a titer maximization problem, where candidates are selected by balancing predicted performance and model uncertainty. The framework is then extended to include probabilistic \ac{CQA} constraints and robustness criteria that account for implementation variability in \acp{CPP}.

    \subsection{Pareto Front Guided Sampling}\label{sec:pfgs}
        
        \ac{PFGS} is used in this work to turn \ac{BO} into a \ac{HitL} decision-support workflow~\cite{stricker_optimal_2024, stricker_pareto_2026}. Instead of returning a single automatically selected optimum, \ac{PFGS} presents a set of candidate experiments on a Pareto front, allowing the expert to inspect the trade-off between expected performance and information gain. In the basic titer maximization example, each candidate recipe is evaluated using the \ac{GP} surrogate through two quantities: the posterior mean titer prediction, $\mu_{\mathrm{Titer}}(\mathbf{x})$, and the corresponding predictive standard deviation, $\sigma_{\mathrm{Titer}}(\mathbf{x})$. The mean prediction represents the expected process performance, while the predictive standard deviation reflects model uncertainty and therefore the potential value of sampling that recipe. Displaying these quantities together makes the exploration--exploitation trade-off explicit and interpretable for the practitioner without the need to choose a fixed hyperparameter to govern the trade-off, as used, e.g., in \ac{UCB}. The workflow is illustrated in Figure~\ref{fig:simple_pareto}.

        \subsubsection{Pareto front}
\begin{figure}[]
            \centering
            \includegraphics[width=0.99\textwidth]{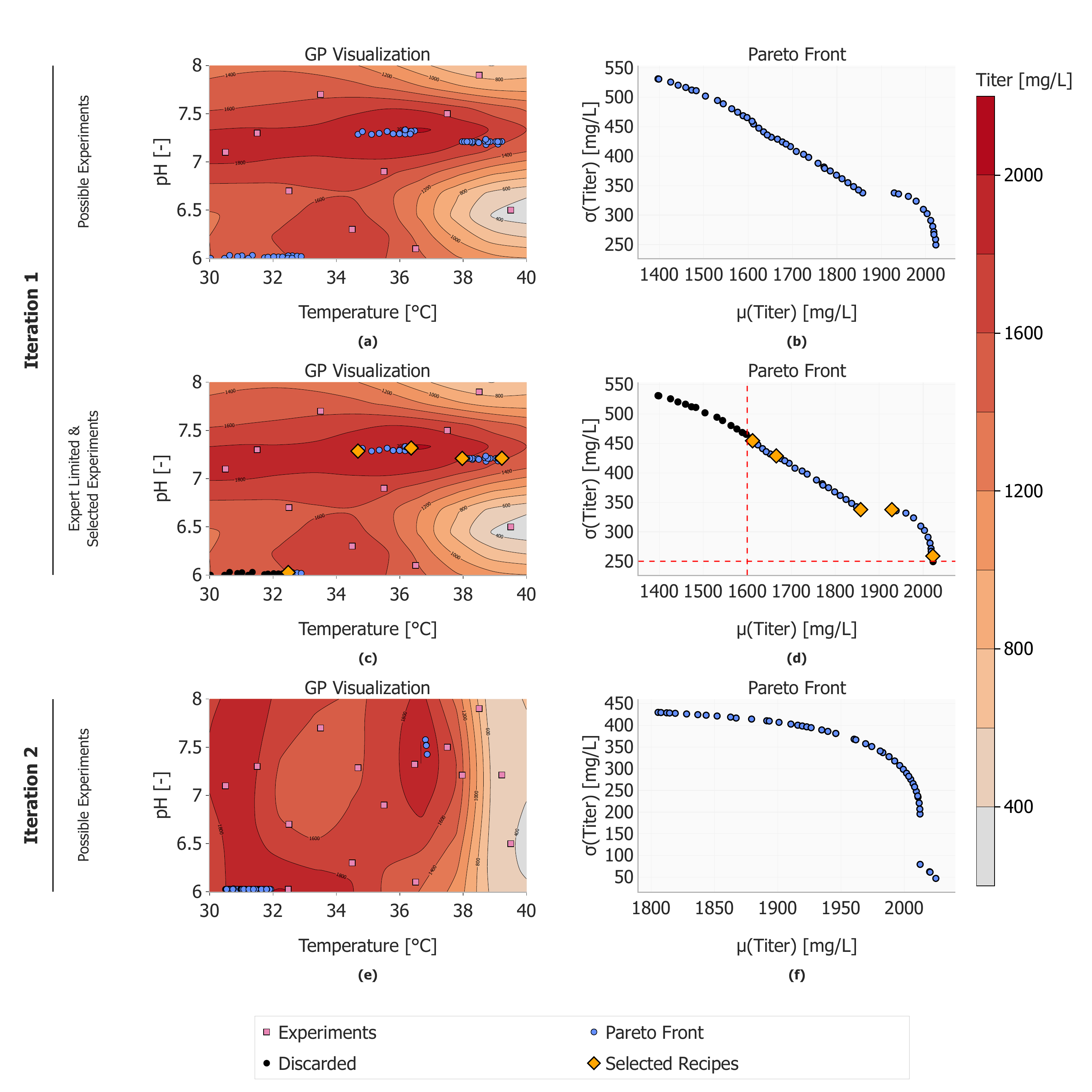}
            \caption{Basic \ac{PFGS} workflow for titer maximization: pink points indicate performed experiments, blue points indicate Pareto-front recipes, black points indicate recipes discarded by the expert, and orange points indicate selected recipes. (a) Initial trained \ac{GP} used to generate the initial Pareto front. (b) Initial Pareto front presenting candidate recipes according to predicted mean titer and predictive standard deviation. (c,d) Expert-defined requirements, such as a minimum expected titer and a minimum uncertainty threshold, restrict both the selectable region of the Pareto front and parts of the investigation space. (e,f) After the selected experiments are evaluated and the \ac{GP} is retrained, the updated Pareto front shows the progress achieved in one iteration, with higher mean titer predictions and lower variance.}
            \label{fig:simple_pareto}
        \end{figure}

        After an initial experimental design has been evaluated, the \ac{GP} is trained and the Pareto front is generated from candidate recipes that are not dominated with respect to predicted titer and predictive uncertainty (Figure~\ref{fig:simple_pareto}~a,b). Candidates with high predicted titer exploit regions that already appear promising, whereas candidates with high uncertainty explore regions that are poorly characterized, far from previous experiments, or associated with strong local variation. Rather than forcing the algorithm to decide between these alternatives automatically, \ac{PFGS} exposes the trade-off to the expert, who can then select experiments according to the current process-development objective.

        \subsubsection{Domain knowledge}
        Expert knowledge is incorporated by specifying requirements on the displayed Pareto objectives. For example, a minimum acceptable predicted titer can be imposed to remove candidates with insufficient expected performance. At the same time, a minimum predictive uncertainty can be required to avoid repeatedly sampling regions that are already well characterized or close to the measurement-noise floor. These requirements are shown in Figure~\ref{fig:simple_pareto}~c,d, where the red vertical line defines the lower bound on predicted titer and the red horizontal line defines the lower bound on predictive uncertainty. Candidate recipes that do not satisfy all requirements are excluded from the selectable region of the Pareto front.

        \subsubsection{Experiment selection}
        From the remaining candidates, the expert can manually select a batch using domain knowledge, practical feasibility considerations, or preferences for diverse operating conditions. Alternatively, the final selection can be delegated to an automated procedure. In this case, all expert-defined requirements are first enforced, after which the feasible point with lowest euclidean distance to the utopia point in normalized objective space is selected. Additional batch points are then chosen to maximize their minimum distance from previously selected candidates and from experiments that have already been performed, thereby encouraging diversity in the investigation space.

        \subsubsection{Active Learning}
        After the selected recipes are evaluated experimentally or in simulation, the new observations are added to the dataset, the \ac{GP} is retrained, and the Pareto front is recomputed. The updated Pareto front in Figure~\ref{fig:simple_pareto}~f can then be compared with the initial front to assess whether the iteration improved the attainable predicted titer, reduced uncertainty in relevant regions, or both. The corresponding change in the \ac{GP} prediction is shown in Figure~\ref{fig:simple_pareto}~e. Based on this updated information, the practitioner can revise the requirements and continue the next \ac{PFGS} iteration. 
        
        \subsubsection{Stopping Criteria} If no candidates remain that satisfy the specified requirements, the expert may either reduce the batch size, relax one or more requirements, or stop the optimization if the region of interest has been characterized to the desired accuracy. The latter can be interpreted as a stopping criterion, indicating that the campaign has reached a point where the desired accuracy has been achieved.

        Once the expert considers the model sufficiently accurate, a final optimization step can be performed that prioritizes both objective maximization and uncertainty minimization. This is particularly relevant for later-stage process development and manufacturing, where the preferred operating point should not only achieve high predicted performance but also be supported by low model uncertainty for regulatory justification.

    \subsection{Constrained Pareto front Guided Sampling} 

\begin{figure}[htbp]
            \centering
            \includegraphics[width=0.55\textwidth]{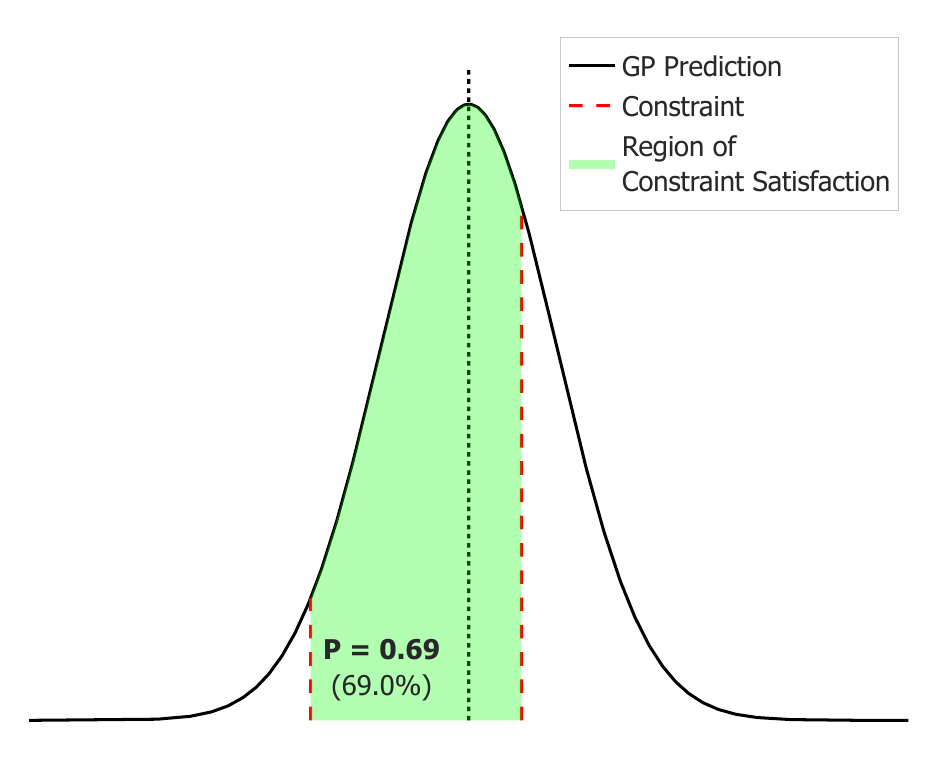}
            \caption{Probabilistic calculation of \ac{CQA} constraint satisfaction for a candidate recipe. The \ac{GP} predictive distribution is shown together with lower and upper specification limits. The shaded area corresponds to $P(\theta_l \leq f_c(\mathbf{x}) \leq \theta_u)$, which is used as a feasibility objective in the constrained \ac{PFGS} formulation.}
            \label{fig:purity_probability}
        \end{figure}

        In the constrained case, \ac{PFGS} is extended by adding the probability of satisfying a  \ac{CQA} as an additional Pareto objective, for example, a purity of a certain level. In general, the process objective may still be maximized, while one or more quality attributes must remain within an acceptable interval. Since the \ac{GP} surrogate provides a predictive distribution for each modeled response, constraint satisfaction can be expressed probabilistically rather than as a deterministic pass--fail criterion. For a candidate recipe $\mathbf{x}$ with \ac{CQA} prediction $f_{c}(\mathbf{x}) \sim \mathcal{N}(\mu_{c}(\mathbf{x}),\sigma^2_{c}(\mathbf{x}))$ and lower and upper specification limits $\theta_l$ and $\theta_u$, the probability of satisfying the two-sided constraint is
        \begin{equation}
            P\!\left(\theta_l \leq f_{c}(\mathbf{x}) \leq \theta_u\right)
            = \Phi\!\left(\frac{\theta_u - \mu_{c}(\mathbf{x})}{\sigma_{c}(\mathbf{x})}\right)
            - \Phi\!\left(\frac{\theta_l - \mu_{c}(\mathbf{x})}{\sigma_{c}(\mathbf{x})}\right),
            \label{eq:purity_constraint}
        \end{equation}
        where $\Phi(\cdot)$ is the standard normal \ac{CDF}. A visualization of this probability calculation is shown in Figure~\ref{fig:purity_probability}. This probability is then treated as an additional Pareto objective alongside the mean titer prediction and the predictive titer uncertainty.

\begin{figure}[htbp]
            \centering
            \includegraphics[width=0.95\textwidth,trim=0 0 18cm 0,clip]{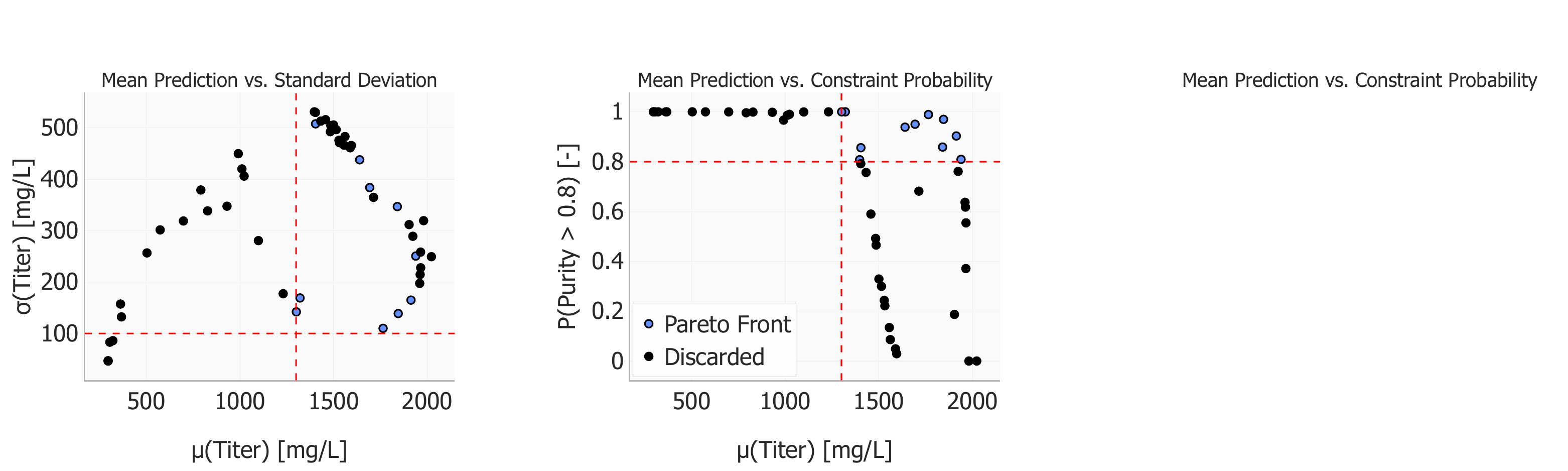}
            \caption{Restricted constrained Pareto front for the first \ac{PFGS} iteration with a purity requirement of $0.8$. The Pareto front was restriced by an expert that expects $\mu_{\mathrm{Titer}}(\mathbf{x})>1300$, $\sigma_{\mathrm{Titer}}(\mathbf{x})>100$ and P$_{\mathrm{Purity>0.8}}(\mathbf{x})>0.8$. (left) Projection of the feasible candidate recipes onto predicted titer and predictive uncertainty, showing the remaining exploration--exploitation trade-off after applying expert requirements. (right) Projection onto predicted titer and probability of purity satisfaction, showing how the \ac{CQA} constraint removes high-titer candidates that are unlikely to meet the required purity threshold. }
            \label{fig:constrained_pareto}
        \end{figure}

        The resulting constrained \ac{PFGS} problem therefore differs from the unconstrained case by replacing the two-dimensional Pareto front with a three-dimensional Pareto front. Candidate recipes are compared with respect to (i) high predicted titer, (ii) high predictive uncertainty or information gain, and (iii) high probability of satisfying the purity constraint. For visualization and expert selection, this three-dimensional Pareto front is shown through two-dimensional projections in Figure~\ref{fig:constrained_pareto}. The first projection relates predicted titer to predictive uncertainty, analogous to the unconstrained case, while the second projection relates predicted titer to the probability of purity satisfaction. Expert requirements can then be applied to all Pareto axes, for example by requiring a minimum expected titer, a minimum uncertainty threshold, and a minimum probability of satisfying purity $>0.8$.

        In the first iteration, these requirements restrict the Pareto front to candidate recipes that are both promising and sufficiently likely to meet the \ac{CQA}. Recipes that fail the purity-confidence requirement are removed even if their predicted titer is high, because they do not satisfy the quality constraint with sufficient probability. The restricted Pareto front therefore provides the expert with a smaller and more relevant set of candidates for batch selection. This highlights the main conceptual difference from the basic \ac{PFGS} workflow: the expert no longer selects only between exploitation and exploration, but also between performance and probabilistic quality assurance.

    \subsection{Robust Pareto Front Guided Sampling}     
        Robustness is incorporated into \ac{PFGS} to identify candidate recipes that remain attractive when the \ac{CPP} values deviate from their nominal set points. This is important because experimental recipes are rarely executed exactly as proposed: pipetting errors, raw-material variability, sensor uncertainty, and scale-dependent disturbances can all shift the realized operating point. In the robust formulation approach, the expert specifies an expected variation (standard deviation) for each \ac{CPP}, and this local uncertainty region is evaluated in addition to the nominal recipe performance. The resulting robustness score can then be added as another Pareto objective, allowing the expert to select candidates that balance high predicted titer, information gain, constraint satisfaction, and stability under \ac{CPP} variation.

        % Two robustness strategies are considered in this work. The first is a worst-case strategy over an expert-defined error box around the candidate recipe. For a proposed recipe $\mathbf{x}$, the error box contains all perturbed recipes $\tilde{\mathbf{x}}$ satisfying $|\tilde{x}_i-x_i|\leq \Delta_i$ for each \ac{CPP} $i$, where $\Delta_i$ denotes the expected implementation tolerance. Robustness can then be quantified by the lowest conservative performance predicted within this box, for example
        % \begin{equation}
        %     \mathcal{R}_{\mathrm{wc}}(\mathbf{x})
        %     = \min_{\tilde{\mathbf{x}}:\, |\tilde{x}_i-x_i|\leq \Delta_i}
        %     \left[\mu(\tilde{\mathbf{x}}) - \beta\sigma(\tilde{\mathbf{x}})\right].
        %     \label{eq:robustness_worst_case}
        % \end{equation}
        % This criterion favors recipes whose lower-confidence performance remains high even under the most adverse admissible perturbation, making it suitable when operating margins are required~\cite{bogunovic_adversarially_2018,paulson_adversarially_2022}, a visual representation is shown in Figure~\ref{fig:robustness_box}.

\begin{figure}
            \centering
            \includegraphics[width=0.5\linewidth]{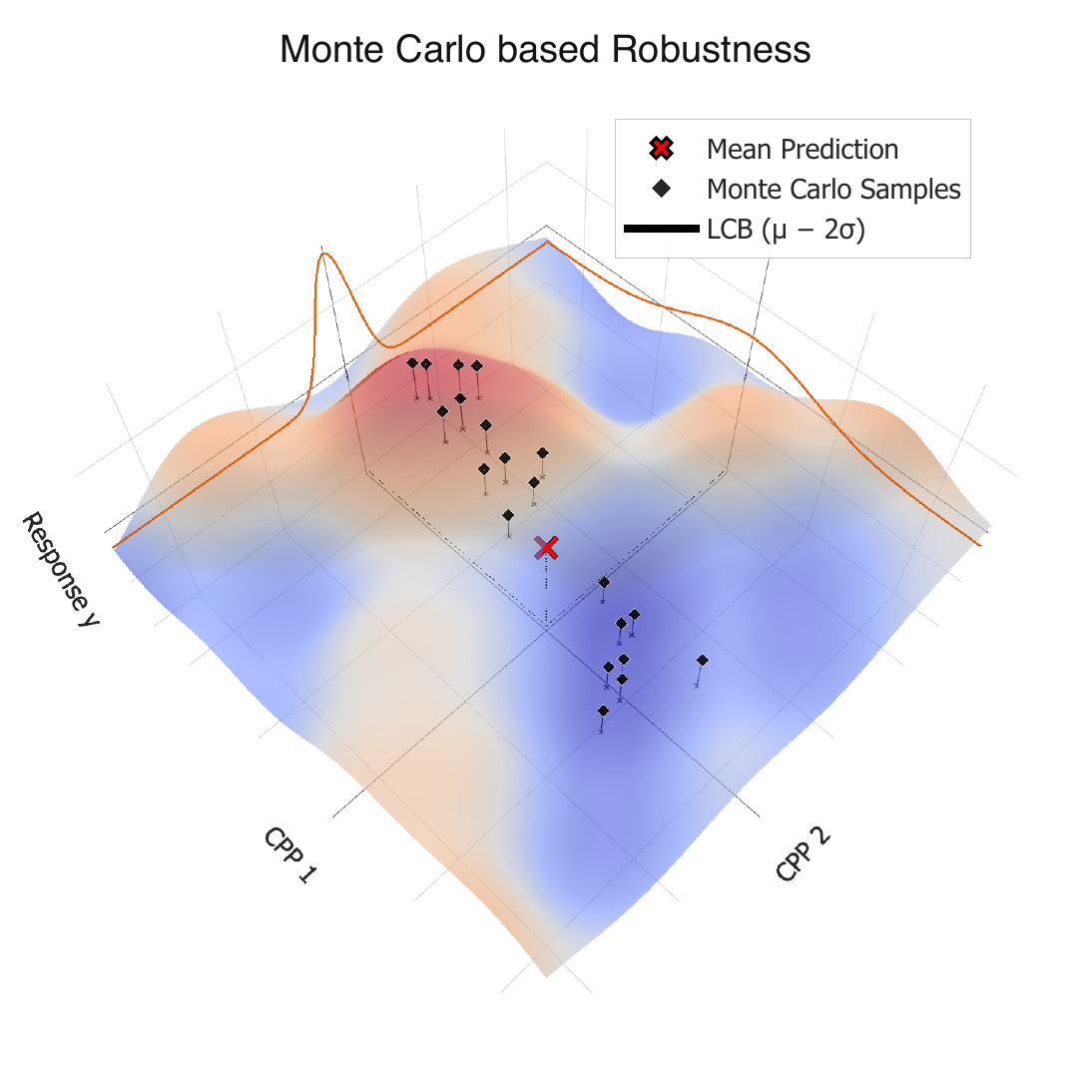}
            \caption{Robustness strategy used to extend \ac{PFGS} beyond nominal recipe optimization. \ac{MC} Sampling-based robustness evaluates the lower-confidence performance over likely \ac{CPP} perturbations around the proposed recipe.}
            \label{fig:robustness_dist}
        \end{figure}

        A \ac{MC} sampling-based strategy was chosen in this work similar to the work of Bogunovic et al.~\cite{bogunovic_adversarially_2018}. It estimates robustness from a distribution of likely \ac{CPP} deviations around the candidate recipe. For each candidate $\mathbf{x}$, $n_s$ perturbed points are sampled from a Gaussian distribution centered at the proposed recipe with expert-defined standard deviations $\sigma_{e,i}$ for each input dimension $i$. The robustness objective is then defined as the mean lower confidence bound of the \ac{GP} titer predictions over these sampled perturbations:
        \begin{equation}
            \mathcal{R}_{\mathrm{MC}}(\mathbf{x}) = \frac{1}{n_s}\sum_{j=1}^{n_s} \left[\mu(\mathbf{x}_j) - \beta \sigma(\mathbf{x}_j)\right], \quad \mathbf{x}_j \sim \mathcal{N}(\mathbf{x},\, \mathrm{diag}(\boldsymbol{\sigma_e}^2)).
            \label{eq:robustness_sampling}
        \end{equation}
        Here, $\mu(\mathbf{x}_j)$ and $\sigma(\mathbf{x}_j)$ are the \ac{GP} posterior mean and standard deviation at perturbation sample $\mathbf{x}_j$, and $\beta = 1.96$ corresponds to a two-sided 95\% confidence interval. Compared with the worst-case strategy, this sampling-based metric, visualized in Figure~\ref{fig:robustness_dist}, reflects average expected robustness over likely deviations rather than the single most conservative point in the uncertainty region.

        % \begin{figure}[htbp]
        %     \centering
        %     \begin{subfigure}[b]{0.48\textwidth}
        %         \centering
        %         \includegraphics[width=\textwidth]{img/visuals/robustness_box.pdf}
        %         \caption{Worst-case robustness}
        %         \label{fig:robustness_box}
        %     \end{subfigure}
        %     \hfill
        %     \begin{subfigure}[b]{0.48\textwidth}
        %         \centering
        %         \includegraphics[width=\textwidth]{img/visuals/robustness_dist.pdf}
        %         \caption{Sampling-based robustness}
        %         \label{fig:robustness_dist}
        %     \end{subfigure}
        %     \caption{Robustness strategies used to extend \ac{PFGS} beyond nominal recipe optimization. (a) Worst-case robustness evaluates the most conservative predicted performance within a bounded \ac{CPP} error box. (b) \ac{MC} Sampling-based robustness evaluates the lower-confidence performance over likely \ac{CPP} perturbations around the proposed recipe.}
        %     \label{fig:robustness_strategies}
        % \end{figure}

\begin{figure}[htbp]
            \centering
            \includegraphics[width=0.99\textwidth]{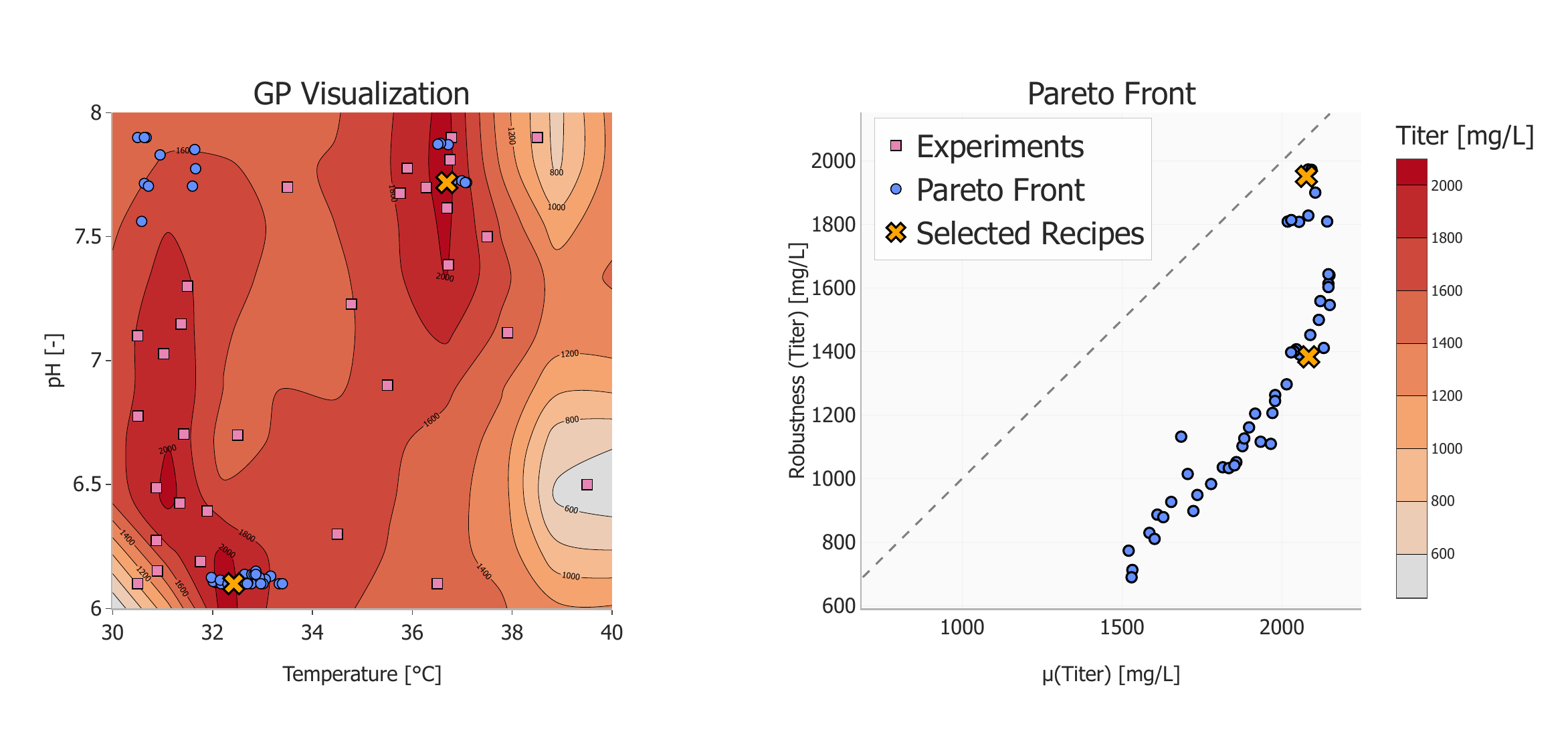}
            \caption{Example of robustness-based candidate interpretation. (right) Pareto representation of nominal predicted titer and sampling-based robust titer. The diagonal reference line indicates the ideal case where robustness equals nominal performance; points closer to this line are less sensitive to the specified \ac{CPP} perturbations. The highlighted candidates have similar nominal performance but different robustness values. (left) Corresponding \ac{GP} prediction illustrating that the more robust candidate lies in a locally stable region with respect to pH, whereas the less robust candidate is located in a region where small pH deviations lead to a stronger performance decrease.}
            \label{fig:robustness_example}
        \end{figure}

       Robustness is introduced as an additional optimization axis and used to support expert decision-making. The robustness criterion must be defined separately for each response of interest, so the robustness of different objectives can be optimized and interpreted independently. A visualization of the resulting Pareto front generated with the sampling method is shown in Figure~\ref{fig:robustness_example}~(right). The diagonal reference line indicates the ideal case in which the robust performance is equal to the nominal prediction. Points close to this line therefore correspond to recipes whose predicted performance does not decrease under the specified \ac{CPP} perturbations. Two candidate recipes are highlighted with orange crosses: one with high robustness and one with lower robustness. When the corresponding \ac{GP} prediction is inspected in Figure~\ref{fig:robustness_example}~(left), the highly robust recipe is located in a region that is insensitive to pH variation, whereas the less robust recipe lies in a region where small pH deviations lead to a stronger performance decrease. Although both candidates have similar nominal performance, their different robustness values are visible on the Pareto front and can be explained by the local shape of the \ac{GP} prediction.

    \subsection{Implementation}
        The implementation, written in Python 3.11, follows a modular structure separating simulation, surrogate modeling, Pareto-front generation, requirement filtering, and candidate selection. \ac{GP} surrogate models were trained for each modeled response on the accumulated experimental dataset with \texttt{BoTorch}~\cite{balandat_botorch_2020}. At each \ac{PFGS} iteration, the surrogate models evaluated a large set of candidate recipes, returning the posterior mean $\mu(\mathbf{x})$ and standard deviation $\sigma(\mathbf{x})$ for each objective. For unconstrained optimization, these quantities were used directly as Pareto objectives. When constraints were present, feasibility was incorporated probabilistically according to Equation~\ref{eq:purity_constraint}. If robust solutions were required, the robustness objective was computed with Equation~\ref{eq:robustness_sampling} and added as an extra Pareto axis. Pareto fronts were generated with the \ac{NSGAII} in \texttt{pymoo}~\cite{blank_pymoo_2020}. A custom Python objective function defined the optimization problem, querying \ac{GP} models trained with \texttt{BoTorch} \cite{balandat_botorch_2020} and returning the required Pareto objectives for each candidate. All relevant hyperparameters are shown in~\ref{appx:hyperparam}. For the \ac{BO} comparison a standard \texttt{BoTorch}~\cite{balandat_botorch_2020} \ac{GP} and acquisition function was used.
        
        After optimization, the resulting non-dominated candidates were filtered according to the expert-defined requirements on the dashboard for manual selection as shown in Figures~\ref{fig:settings}--\ref{fig:select} or processed by the automated selection routine described above.

         The overall algorithmic workflow is summarized in Algorithm~\ref{alg:pfgs}. The same sequence is used for the unconstrained, constrained, and robust variants; only the objective vector $\mathbf{f}_t(\mathbf{x})$ and the expert-defined requirements $\mathcal{R}$ change depending on the problem formulation.

\definecolor{commentgray}{RGB}{110,110,110}
\definecolor{commentgray}{RGB}{130,130,130}
\definecolor{guidegray}{RGB}{175,175,175}
% every algorithm row is forced to the same pitch (\algrh above the baseline,
% \algrd below) so the indent guides below can tile without gaps
\newcommand{\algrh}{1.30\baselineskip}
\newcommand{\algrd}{0.45\baselineskip}
\newcommand{\rstrut}{\rule[-\algrd]{0pt}{\dimexpr\algrh+\algrd\relax}}
\newcolumntype{N}{>{\rstrut\bfseries}p{0.042\textwidth}}
% pseudocode keywords: bold in the monospace font (needs \ttdefault=lmtt, set in main.tex)
\newcommand{\kw}[1]{\textbf{#1}}
\newcolumntype{M}{>{\raggedright\arraybackslash}p{0.55\textwidth}}
% comments: an l cell, not a p box, so they share the row baseline and add no height
\newcolumntype{K}{>{\color{commentgray}\itshape}l}
\newcommand{\cmt}[1]{\textcolor{commentgray}{\ttfamily\itshape // #1}}
% --- indentation guides: one vertical rule per open block level ---
\newlength{\indw}\setlength{\indw}{1.0em}
% the rule is smashed (zero height/depth) so it never changes the row height, and
% it spans exactly one row pitch, so the rules of consecutive rows meet edge to edge
\newcommand{\gbar}{\smash{\textcolor{guidegray}{\rule[-\algrd]{0.4pt}{\dimexpr\algrh+\algrd\relax}}}}
\newcommand{\indA}{\leavevmode\hspace*{0.3em}\gbar\hspace*{\indw}}
\newcommand{\indB}{\indA\gbar\hspace*{\indw}}
\newcommand{\indC}{\indB\gbar\hspace*{\indw}}

\begin{algorithm}[htbp]
    \caption{Human-in-the-Loop Pareto Front Guided Sampling (\ac{PFGS}).}
    \label{alg:pfgs}
    \small\ttfamily\mathversion{algott}
    \setlength{\extrarowheight}{0pt}
    \renewcommand{\arraystretch}{1}
    \begin{tabularx}{\linewidth}{N >{\raggedright\arraybackslash}X K}
        \toprule
        & \kw{Input:}\ $\mathcal{D}_0,\; q$ & \cmt{dataset, batch size} \\
        \midrule
        1  & \kw{while} experimental budget not exhausted \kw{do:} & \\
        2  & \indA\kw{fit} $\mathcal{GP}(\mathbf{x})$ on $\mathcal{D}_t = \{(\mathbf{x}_i, \mathbf{y}_i)\}_{i=1}^{n_d}$ & \\
        3  & \indA\kw{init} \ac{NSGAII} population $\mathbf{X}^{(0)}$ & \\
        4  & \indA\kw{while} not converged \kw{do:} & \\
        5  & \indB$\hat{\mathbf{y}},\,\hat{\boldsymbol{\sigma}} \leftarrow \mathcal{GP}(\mathbf{X}^{(g)})$ & \cmt{\ac{GP} posterior} \\
        6  & \indB$\mathbf{f}_t(\mathbf{x}) \leftarrow \bigl[\mu_t,\;\sigma_t,\;P_t(\theta_l \leq f_c \leq \theta_u),\;\mathcal{R}_t\bigr]$ & \cmt{4-obj. fitness} \\
        7  & \indB\kw{select} via non-dom.\ sort $+$ crowding dist. & \cmt{\ac{NSGAII} ranking} \\
        8  & \indB\kw{check} convergence ($\Delta\text{HV} < \varepsilon$ or max gen.) & \cmt{stopping criterion} \\
        9  & \indB\kw{apply} crossover \& mutation $\to \mathbf{X}^{(g+1)}$ & \cmt{genetic operators} \\
        10 & \indA\kw{return} $\mathbf{X}^{*}_t \leftarrow$ Pareto front of $\mathbf{f}_t$ & \cmt{non-dominated set} \\
        11 & \indA\kw{visualize} $\mathbf{X}^{*}_t$ on interactive dashboard & \cmt{human-in-the-loop} \\
        12 & \indA\kw{get} expert thresholds $\boldsymbol{\tau}_{\mathcal{R}}$ on $\mathbf{f}_t$ & \cmt{expert input} \\
        13 & \indA$\mathcal{P}_{\mathcal{R},t} \leftarrow \{\mathbf{x} \in \mathbf{X}^{*}_t : \mathbf{f}_t(\mathbf{x}) \geq \boldsymbol{\tau}_{\mathcal{R}}\}$ & \cmt{feasible part of Pareto front} \\
        14 & \indA\kw{if} $\mathcal{P}_{\mathcal{R},t} = \emptyset$\kw{:} & \cmt{stopping criteria} \\
        15 & \indB\kw{stop} & \\
        16 & \indA\kw{select} $\mathcal{B}_t \subseteq \mathcal{P}_{\mathcal{R},t}$,\; $|\mathcal{B}_t| \leq q$ & \cmt{automated or manual selection} \\
        17 & \indA\kw{run} experiments $\to \mathbf{Y}_t = \{\mathbf{y}(\mathbf{x}) : \mathbf{x} \in \mathcal{B}_t\}$ & \cmt{run selected experiments} \\
        18 & \indA$\mathcal{D}_{t+1} \leftarrow \mathcal{D}_t \cup \{(\mathbf{x},\,\mathbf{y}(\mathbf{x})) : \mathbf{x} \in \mathcal{B}_t\}$ & \cmt{new dataset} \\
        19 & \indA\kw{increment} $t \leftarrow t + 1$ & \cmt{next iteration} \\
        \bottomrule
    \end{tabularx}
\end{algorithm}
        
        The code will be made available on the Autonomous Industrial Systems Lab GitHub page at \url{https://github.com/AISL-at-Imperial-College-London}. 
        
        Generative AI tools (Claude Sonnet 5, Claude Opus 5) were used to assist with coding, and all content was reviewed and approved by the authors.

%% file: 02.1_CaseStudies.tex
\section{Case Study}
\label{sec:case_study}
\begin{figure}[!ht]
            \centering
            \includegraphics[width=0.75\textwidth]{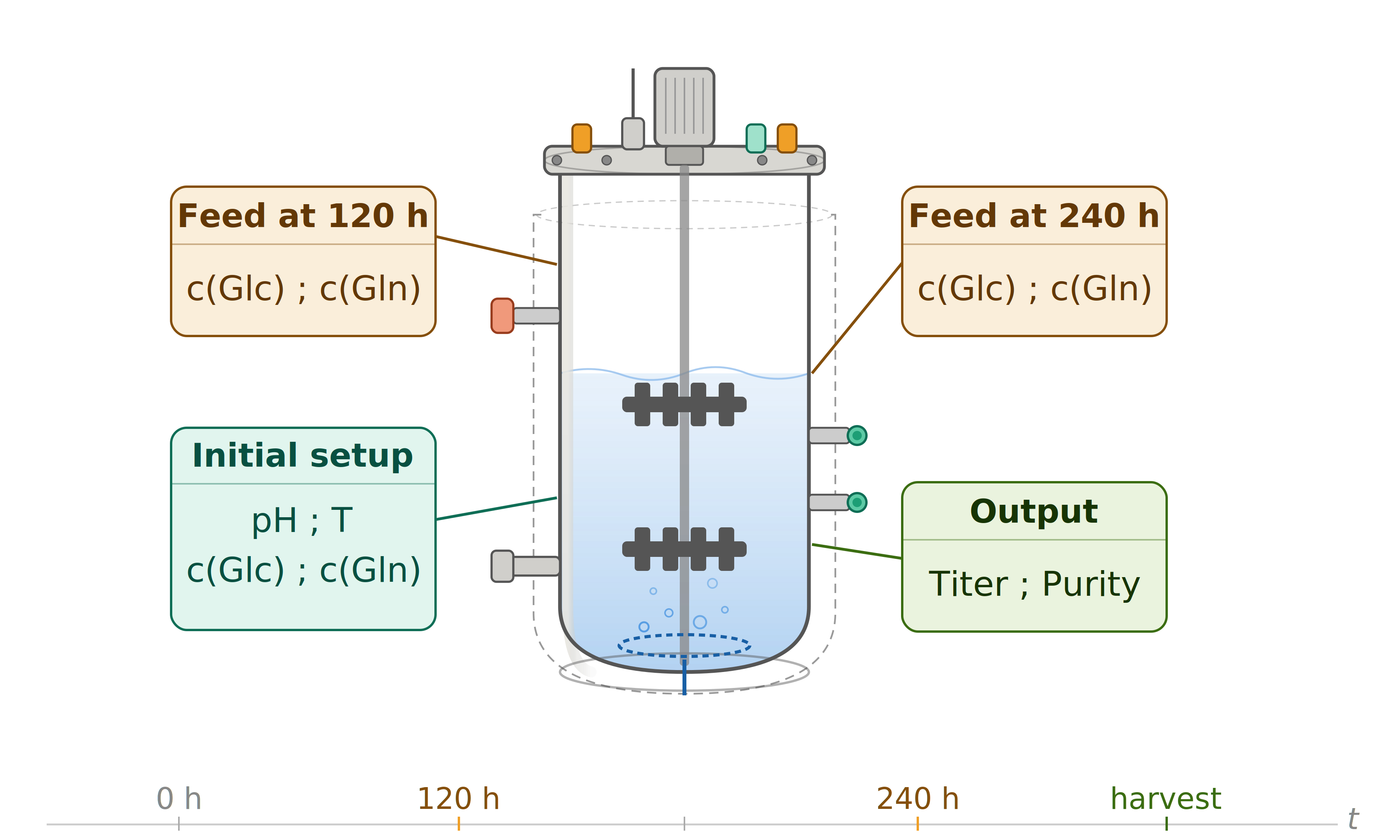}
            \caption{Schematic of the bioreactor process, including the initial setup, two bolus feeds, and the final harvest.}
            \label{fig:bioreactor}
        \end{figure}

    The bioprocess case study uses an eight-dimensional, three-output fed-batch simulator described by a system of ordinary differential equations shown in~\ref{app:bio-simulator}. The model structure is based on the formulations reported by \cite{craven_process_2013,martens_holistic_2025} and is extended here to include aggregation and purity alongside titer as process outputs. The simulated process represents a two-week fed-batch reactor operation with two scheduled feeding events on days~5 and~10 (Figure~\ref{fig:bioreactor}). The decision variables comprise the initial culture conditions and feed concentrations for glucose and glutamine, while the outputs describe product formation and quality-related attributes. Table~\ref{tab:simulator_params} summarizes the input ranges and model outputs used throughout the case study. Within this work, the simulator provides a controlled yet bioprocess-relevant benchmark for evaluating optimization performance in a high-dimensional setting where direct visualization is limited and expert interpretability becomes increasingly important.

    \begin{table}[h]
            \centering
            \caption{Input variables and output responses of the fed-batch bioprocess simulator.}
            \label{tab:simulator_params}
            \vspace{0.2cm}
            \begin{tabular}{lccccl@{\hskip 4em}lcc}
                \toprule
                \multicolumn{5}{c}{\textbf{Inputs}} && \multicolumn{3}{c}{\textbf{Outputs}} \\
                \cmidrule(r{1em}){1-5} \cmidrule(l){7-9}
                \textbf{Name} & \textbf{Range} & \textbf{Unit} & \textbf{Standard Deviation} & \textbf{Default} && \textbf{Name} & \textbf{Range}  & \textbf{Unit} \\
                \cmidrule(r{1em}){1-5} \cmidrule(l){7-9}
                Temperature  & {[}30.0,\ 40.0{]}   & $^\circ C$  &0.1&33.4&& Titer       & --  & g/L \\
                pH           & {[}6.0,\ 8.0{]}   & --  &0.2 &6.7&& Aggregation & --  & \%  \\
                Initial GLC  & {[}0.0,\ 4.0{]}  & g/L &0.05&4.0&& Purity      & --  & \%  \\
                Initial GLN  & {[}0.0,\ 4.0{]}   & g/L &0.05&4.0&&             &     &     \\
                Feed 1 GLC   & {[}0.0,\ 4.0{]} & g/L &0.05&2.0&&             &     &     \\
                Feed 1 GLN   & {[}0.0,\ 4.0{]}  & g/L &0.05&1.0&&             &     &     \\
                Feed 2 GLC   & {[}0.0,\ 4.0{]} & g/L &0.05&2.0&&             &     &     \\
                Feed 2 GLN   & {[}0.0,\ 4.0{]}  & g/L &0.05&1.0&&             &     &     \\
                \bottomrule
            \end{tabular}
        \end{table}
        
        The optimization problem is formulated using endpoint data at the end of the 14-day cultivation period; therefore, the sequential decision-making framework does not explicitly model the full process dynamics. However, the \ac{PFGS} methodology is not limited to static endpoint models. As long as the predictive model provides a point and associated uncertainty estimate for each candidate condition, the same framework can also be applied to time-resolved process models.

\begin{figure}[ht]
            \centering
            \begin{subfigure}[b]{0.32\textwidth}
                \centering
                \includegraphics[width=\textwidth]{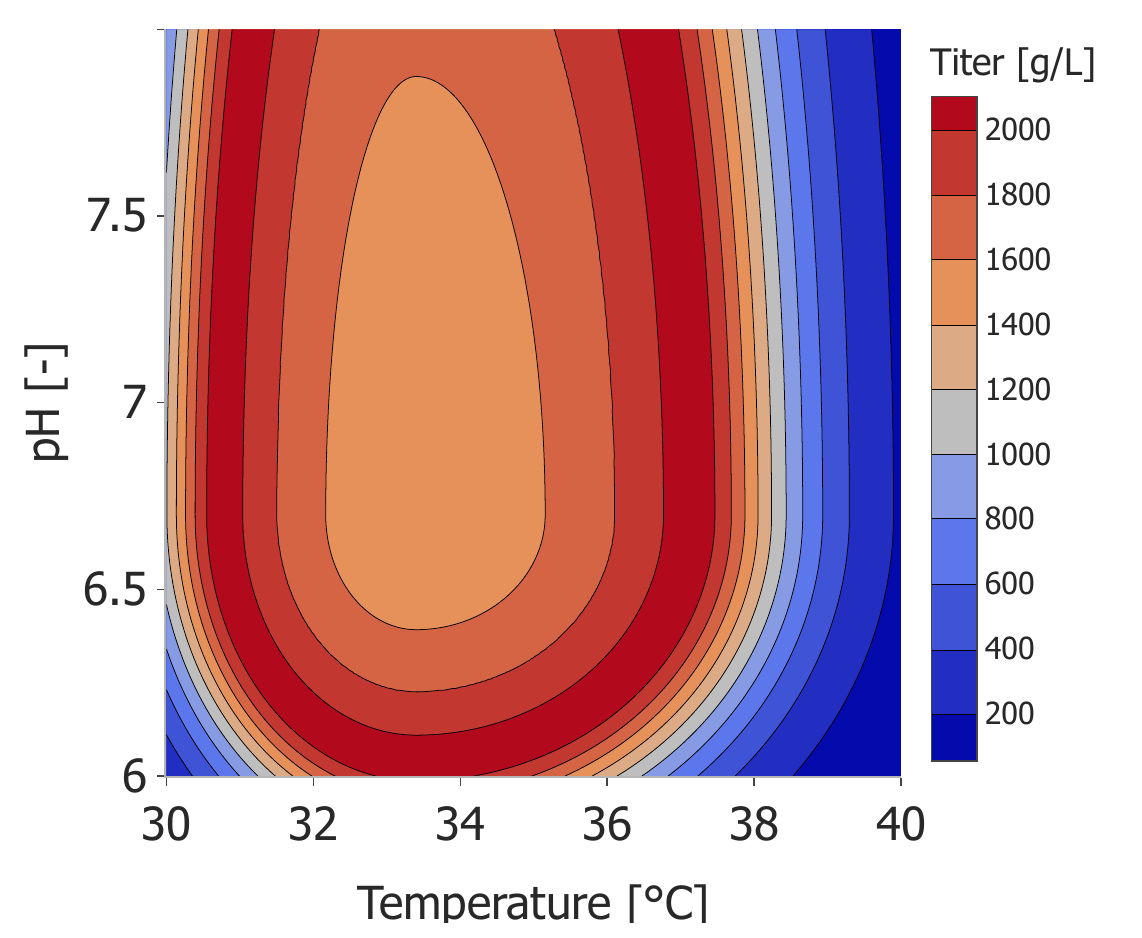}
                \caption{Titer objective}
                \label{fig:bio_titer_landscape}
            \end{subfigure}
            \hfill
            \begin{subfigure}[b]{0.32\textwidth}
                \centering
                \includegraphics[width=\textwidth]{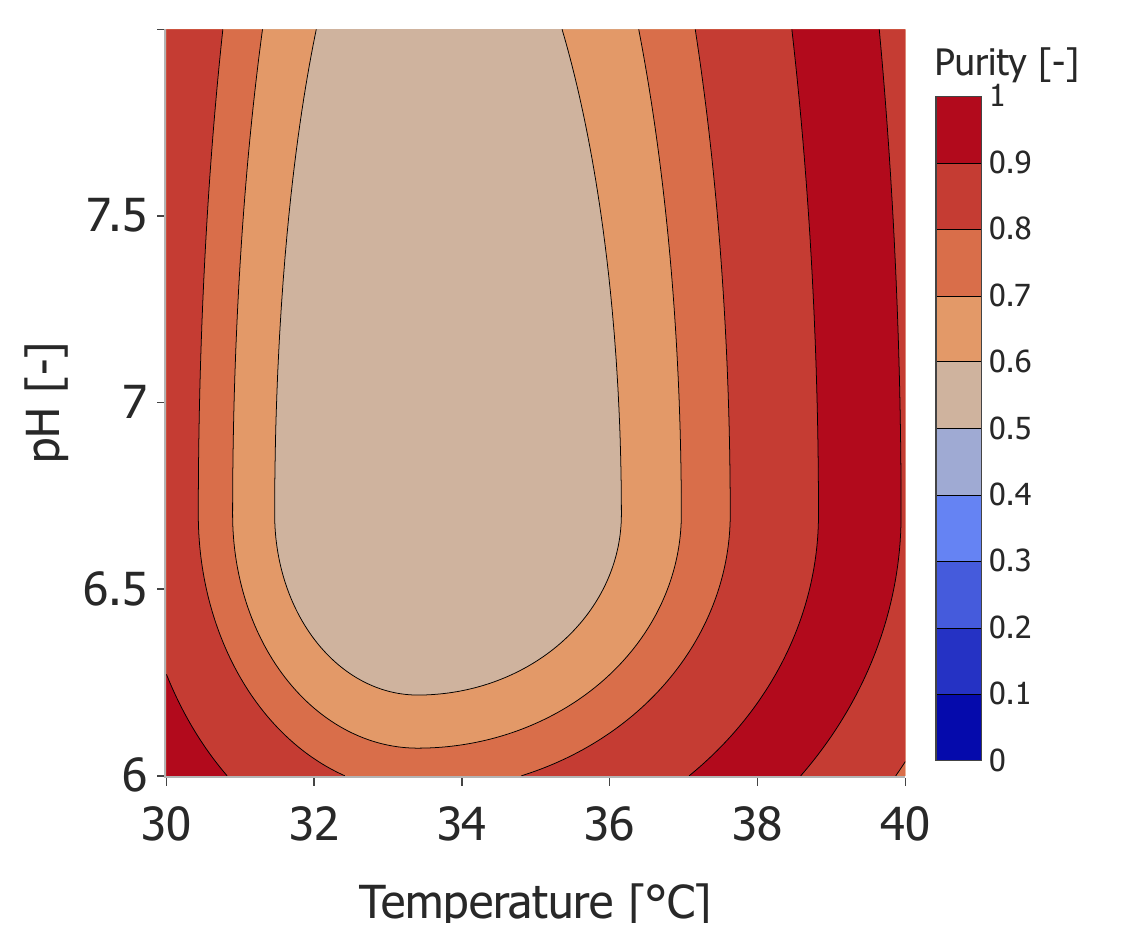}
                \caption{Purity response}
                \label{fig:bio_purity_landscape}
            \end{subfigure}
            \hfill
            \begin{subfigure}[b]{0.32\textwidth}
                \centering
                \includegraphics[width=\textwidth]{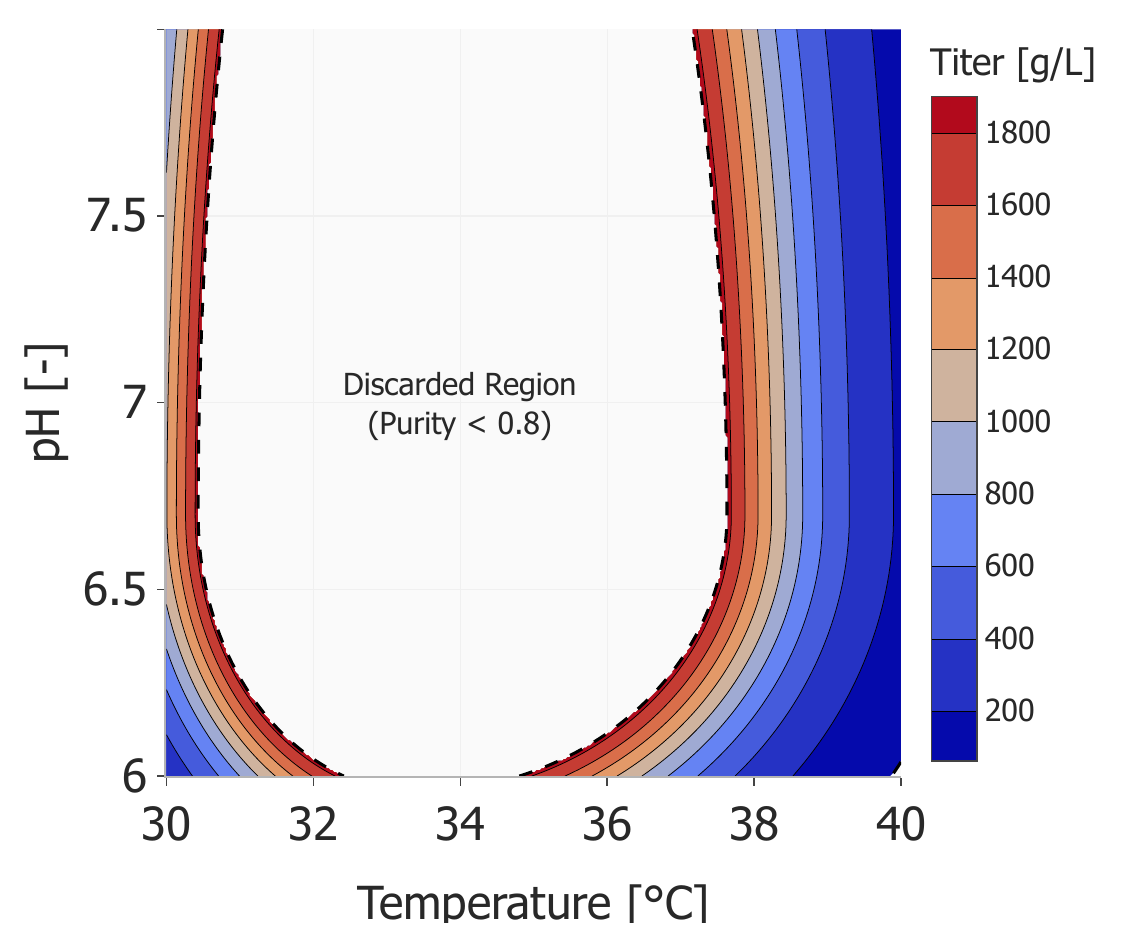}
                \caption{Constrained objective}
                \label{fig:bio_constrained_landscape}
            \end{subfigure}
            \caption{Two-dimensional slice of the bioprocess simulator with all non-displayed inputs fixed at their default values. (a) Titer response surface used as the optimization objective. (b) Purity response surface used to define the quality constraint. (c) Resulting constrained optimization landscape for titer maximization subject to purity $>0.8$ (black dashed line), showing how the feasible region restricts the objective search space.}
            \label{fig:bio_landscape_2d}
        \end{figure}

        To provide an interpretable visualization of the eight-dimensional case study, a two-dimensional version of the simulator is used to showcase some effects while the remaining six inputs are fixed at the default values listed in Table~\ref{tab:simulator_params}. This two dimensional problem is shown in Figure~\ref{fig:bio_titer_landscape}  for the titer objective, the purity response (Figure~\ref{fig:bio_purity_landscape}), and the constrained optimization problem obtained by requiring purity to be greater than 0.8 (Figure~\ref{fig:bio_constrained_landscape}).
        
%% file: 03_results.tex
\section{Results \& Discussion}
\label{sec:results}
   This section evaluates the proposed \ac{HitL} \ac{PFGS} workflow on the bioprocess case study described in Section~\ref{sec:case_study}. First, the optimization problem is defined, including the Pareto objectives used to represent performance, uncertainty, constraint satisfaction, and robustness. The subsequent sections discuss how these Pareto-front projections can support expert decision-making during the optimization campaign and assess the improvement achieved over successive iterations.

   \subsection{Setup of the Optimization Problem}\label{sec:results_setup}
      The optimization objective was setup to identify recipes with high predicted titer while satisfying a minimum purity requirement of approximately $80\%$. This intentionally broad specification reflects the flexibility of \ac{PFGS}, which allows objectives and preferences to be adapted throughout the optimization campaign rather than fixed \textit{a priori}. As a result, decision-making criteria can evolve in response to newly acquired process knowledge and changing development priorities. In addition, robustness is considered by accounting for expected implementation variability in the input variables. These perturbations are modeled using input-specific distributions with standard deviations defined for the investigated \acp{CPP} listed in Table~\ref{tab:simulator_params}. These values are specified by the operator, who often has practical knowledge of the expected \ac{CPP} variation from previous processes, alternatively, they could be estimated from historical process data. 

      Following the standard \ac{PFGS} formulation described in Section~\ref{sec:pfgs}, titer optimization is represented by two Pareto objectives: the \ac{GP} posterior mean titer, $\mu_{\mathrm{Titer}}(\mathbf{x})$, and the corresponding predictive standard deviation, $\sigma_{\mathrm{Titer}}(\mathbf{x})$. The mean prediction captures expected process performance, whereas the predictive standard deviation represents model uncertainty and can be seen as a proxy for information gain from sampling a candidate recipe.

      To incorporate the purity constraint, the probability of satisfying the required purity threshold is added as an additional Pareto objective, following Equation~\ref{eq:purity_constraint}. Robustness is included using the sampling-based formulation in Equation~\ref{eq:robustness_sampling}, using the estimated expected lower-confidence titer concentration under likely \ac{CPP} deviations. The resulting optimization problem therefore uses a four-dimensional Pareto representation consisting of predicted titer, titer uncertainty, probability of purity satisfaction, and robust titer performance.

\begin{figure}[ht!]
       \centering
       \begin{subfigure}{\textwidth}
          \centering
          \includegraphics[width=0.99\textwidth,trim=0 0 0 0,clip]{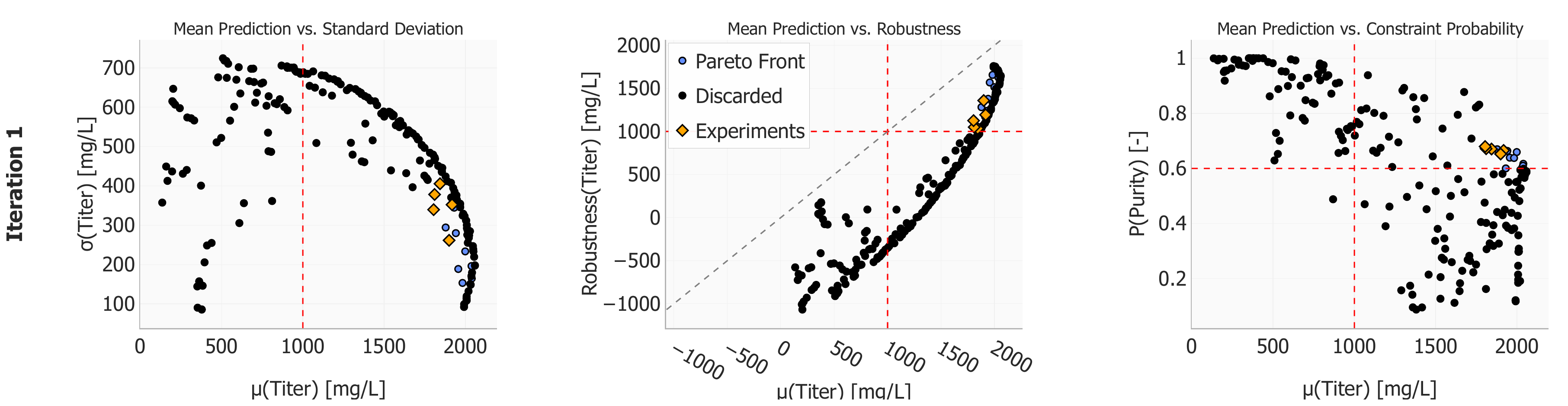}
          \caption{}
          \label{fig:res_initial_pareto}
       \end{subfigure}
    
       \vspace{1em}
    
       \begin{subfigure}{\textwidth}
          \centering
          \includegraphics[width=0.99\textwidth,trim=0 0 0 0,clip]{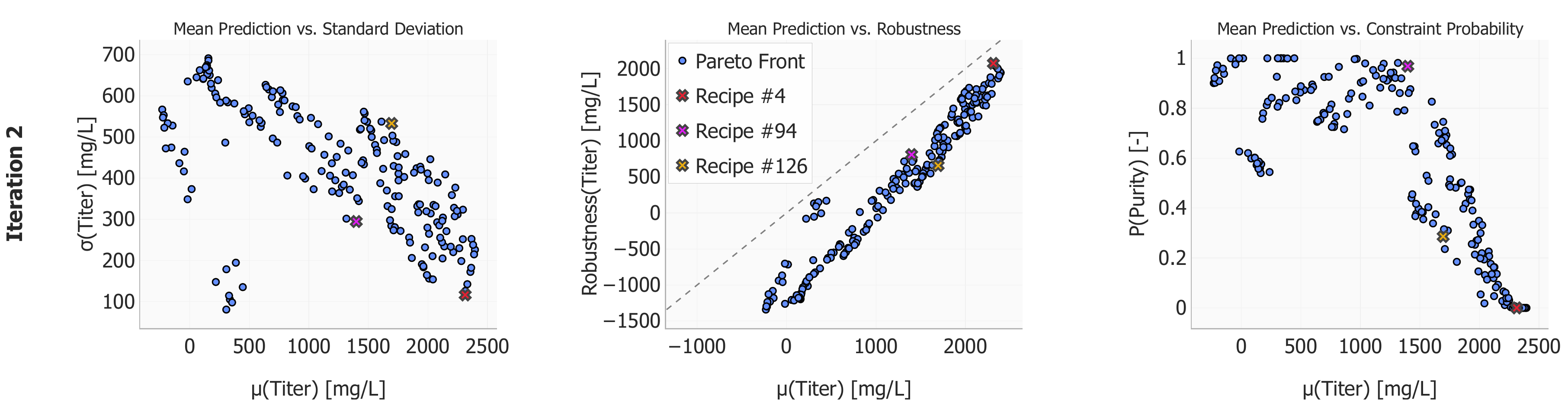}
          \caption{}
          \label{fig:res_mid_pareto}
       \end{subfigure}
    
       \vspace{1em}
    
       \begin{subfigure}{\textwidth}
          \centering
          \includegraphics[width=0.99\textwidth,trim=0 0 0 0,clip]{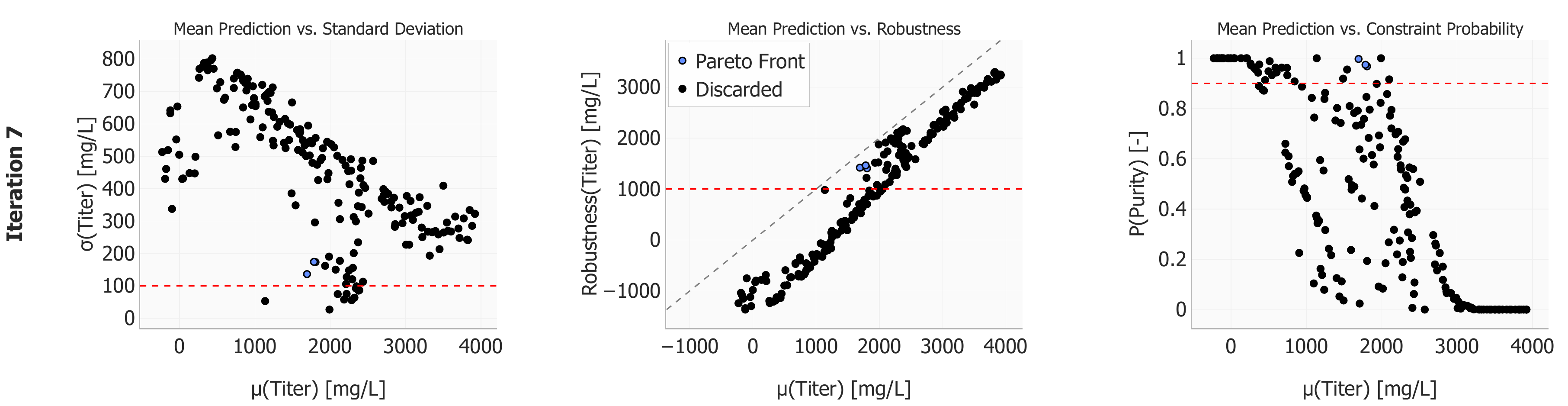}
          \caption{}
          \label{fig:res_late_pareto}
       \end{subfigure}
        \caption{Evolution of the Pareto front throughout the optimization campaign. Blue points denote Pareto-optimal candidates, while black points represent recipes rejected by the expert. Furthermore, orange diamonds indicate the recipes selected by the expert. Each row shows three projections: (left) the classical \ac{PFGS} projection of the \ac{GP} posterior mean titer, $\mu_{\mathrm{Titer}}(\mathbf{x})$, versus the predictive standard deviation, $\sigma_{\mathrm{Titer}}(\mathbf{x})$; (middle) the robustness projection for the titer prediction, where candidates on or above the diagonal are expected to retain their nominal performance under the specified \ac{CPP} perturbations; and (right) the probability of satisfying the purity constraint, $P(\mathrm{Purity}>0.8)$, plotted against the \ac{GP} posterior mean titer. (a) shows the initial Pareto front used to select experiments for the subsequent optimization iteration, resulting in the updated Pareto front shown in (b). Panel (c) depicts the Pareto front at a later stage of the campaign, after six \ac{PFGS} iterations.}
       \label{fig:res_pareto_evolution}
    \end{figure}

      The optimization campaign is initialized with 10 experiments sampled using \ac{LHS}. Separate \ac{GP} surrogate models are then trained for titer and purity using the initial dataset. The \ac{NSGAII} algorithm was used to create the four-dimensional Pareto fronts and it was checked for convergence. The Pareto front after the initial sampling stage shown in Figure~\ref{fig:res_initial_pareto} provides the first set of candidate recipes for expert inspection and selection.

   \subsection{Pareto-Front Analysis}\label{sec:pareto_analysis}
      The multi-dimensional Pareto front is shown as 2D projections, which is still hard to visualize but the interactive dashboard shown in~\ref{app:dashboard} allows operators to investigate a particular recipe and see its position highlighted in all projections as well as in the parallel coordinate plot. Form these the expert can define practical selection requirements. The first criterion considered here is the probability of satisfying the \ac{CQA} constraint, shown in Figure~\ref{fig:res_initial_pareto} (right), because purity is a critical feasibility requirement. At this early stage of development, the expert observes that achieving a high probability of $\mathrm{Purity}>0.8$ is challenging across the candidate set. Therefore, a relatively permissive minimum threshold of $60\%$ is selected for the probability of purity satisfaction.

      The expert then considers the \ac{GP} posterior mean titer, $\mu_{\mathrm{Titer}}(\mathbf{x})$, shown on the horizontal axes of the Pareto-front projections. Since the initial Pareto front contains predicted titers up to approximately 2000~mg/L, candidates with predicted titer below 1000~mg/L are excluded from further consideration. These low-performance regions are not prioritized because they are unlikely to contribute useful process-development information relative to higher-performing alternatives.

      Robustness is evaluated next using the robust titer projection in Figure~\ref{fig:res_initial_pareto} (middle). The expert again selects a threshold of 1000~mg/L, requiring candidates with lower nominal predicted titer to at least maintain acceptable performance under the specified \ac{CPP} perturbations. This prevents the selection of recipes whose already moderate predicted performance would decrease further under expected input variability. Finally, the remaining candidates are inspected with respect to predictive uncertainty. Since all remaining Pareto candidates have $\sigma_{\mathrm{Titer}}(\mathbf{x})>100$, no additional requirement is imposed on this axis.

      After applying these requirements, the expert selects a batch of five experiments. Because the campaign is still in an early development stage, the selected recipes are chosen from the remaining candidates with the highest predictive standard deviation. This prioritizes information gain and supports rapid improvement of the surrogate model. The selected experiments are then evaluated experimentally or in simulation, the new observations are added to the dataset, and the \ac{GP} models are retrained before generating the next Pareto front.

      The Pareto front obtained after the second iteration (Figure~\ref{fig:res_mid_pareto}) shows a clear improvement in the explored design space. In the classical \ac{PFGS} projection, the reachable predicted titer range increases, indicating that the surrogate model has identified more promising high-performance regions. The candidates in the robustness projection move closer to the diagonal, suggesting that a larger fraction retains their nominal predicted performance under the specified \ac{CPP} perturbations. In addition, higher predicted titers can now be achieved with larger probabilities of satisfying the purity constraint. Three representative candidate recipes are highlighted with crosses: the red candidate combines high predicted titer and robustness but has a low probability of satisfying the purity constraint; the pink candidate offers a high probability of constraint satisfaction with moderate titer and robustness but lower expected information gain; and the orange candidate provides a favorable trade-off between performance and information gain, although with comparatively lower robustness and constraint satisfaction probability. This highlights both the complexity of expert decision-making and the process insight that can be gained by interpreting the Pareto-front trade-offs.

\begin{figure}[!ht]
            \centering
            \includegraphics[width=0.85\textwidth]{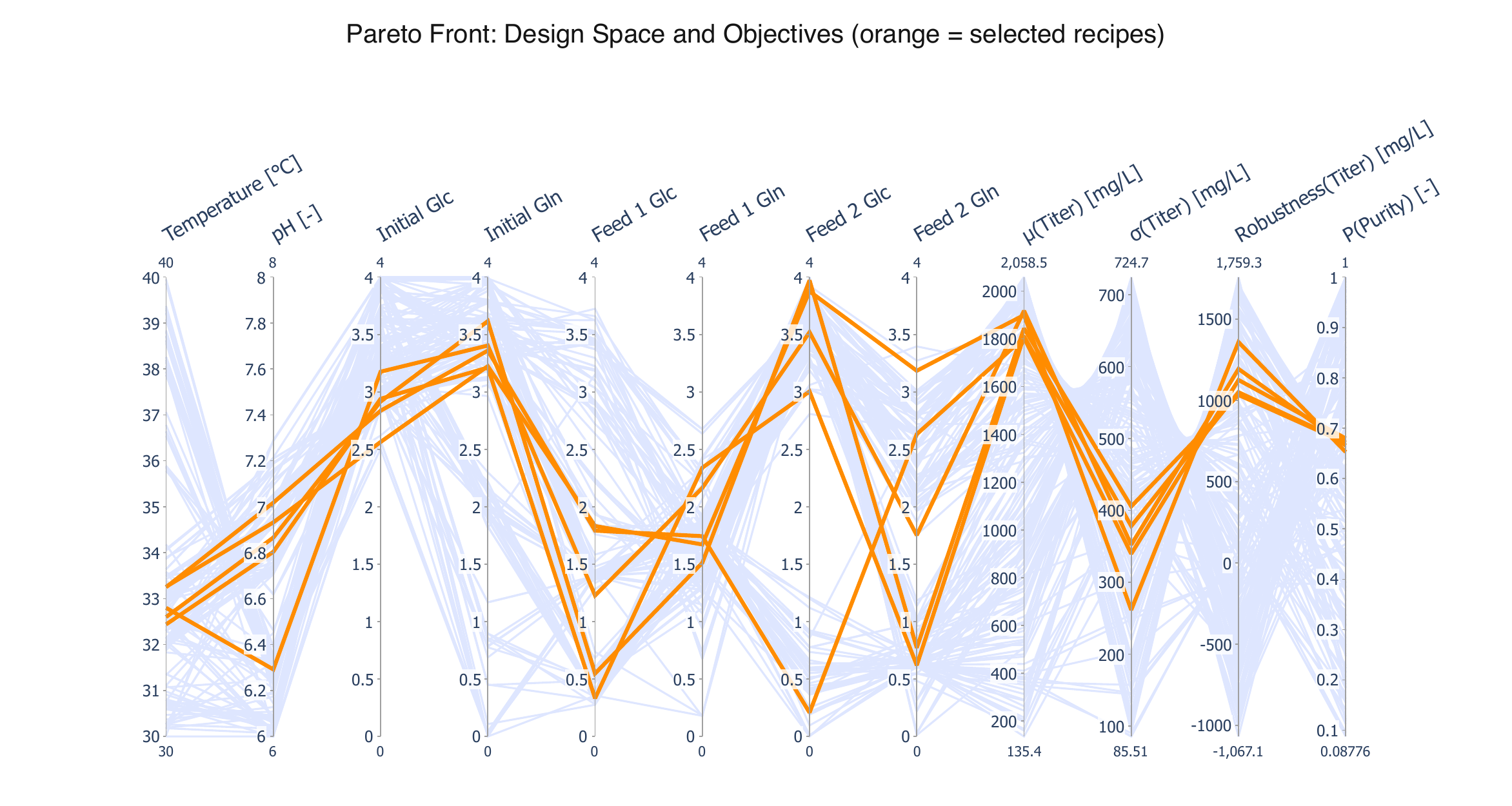}
            \caption{Parallel coordinate plot of the Pareto front. The eight decision variables are shown on the left, followed by the objectives; the selected recipes are highlighted in orange.}
            \label{fig:parallel}
        \end{figure}

      Another way to visualize the Pareto front for the expert user is with a parallel coordinate plot, shown in Figure~\ref{fig:parallel}. Here, the eight decision variables are shown on the left, followed by the objectives, for all recipes on the Pareto front. In this view the user can, for example, highlight regions of interest and see where they lie in the investigation space. The selected recipes are highlighted in orange.

   \subsection{Late-Stage Pareto Front}
      At a later stage of the development campaign, after six \ac{PFGS} iterations have been performed, the progress of the optimization is visible in the updated Pareto front shown in Figure~\ref{fig:res_late_pareto}. Compared with the initial Pareto front in Figure~\ref{fig:res_initial_pareto}, the predicted titer values extend to substantially higher levels, indicating improved model knowledge of high-performing regions. The robustness projection also exhibits a more diagonal structure, suggesting that a larger fraction of candidates retain their nominal predicted performance under the specified \ac{CPP} perturbations. In addition, the probability of satisfying the purity requirement approaches 100\% for candidates with predicted titers up to approximately 2000~mg/L, whereas in the initial iteration high feasibility probabilities were primarily observed at lower titer values. These changes illustrate that \ac{PFGS} provides not only candidate recipes for the next iteration, but also interpretable insight into the current state of process understanding and the performance levels that can realistically be expected.

      Because the campaign is now at a later development stage, the expert applies more conservative selection requirements. The minimum probability of satisfying the purity constraint is increased to 90\%, while the robust titer threshold is maintained at 1000~mg/L. The remaining candidates are then inspected with respect to predictive uncertainty. Candidates with $\sigma_{\mathrm{Titer}}(\mathbf{x})<100$~mg/L are discarded because their uncertainty is close to the expected experimental variation in titer and therefore provides limited additional information gain. After applying these requirements, only three candidate recipes remain. The expert therefore reduces the experimental batch from five to three reactors to conserve resources. Alternatively, the unused reactor capacity could be used to generate replicates of promising operating conditions and estimate experimental repeatability.

      This example highlights how expert expectations and selection criteria can evolve during process development. Regions discarded in one iteration may differ from those discarded later as the surrogate model becomes more informative and the development objective shifts from broad exploration toward more conservative refinement. This adaptability is a central advantage of \ac{PFGS}: instead of fixing a single acquisition function or objective weighting a priori, the expert can decide which objective should be prioritized in the current iteration based on the visualized Pareto-front trade-offs.

    \subsection{Generalization of \acs{PFGS}}
        The example above demonstrates how \ac{PFGS} can be used for constrained and robust optimization. More generally, however, \ac{PFGS} is not limited to the specific objectives used in this case study. A wide range of \ac{BO} acquisition functions or surrogate-derived quantities can be reformulated as Pareto objectives and presented in a \ac{HitL} decision framework. This allows operators and domain scientists to inspect the model recommendations, compare competing objectives, and incorporate process knowledge before selecting the next experiments. The framework therefore does not introduce a new acquisition function, but reorganizes existing \ac{BO} objectives into an interpretable multi-objective format that supports expert decision-making and the integration of domain knowledge. A visualization of the interactive dashboard used to include the human into the loop is shown in~\ref{app:dashboard}.

    \subsection{Study of Expert Personas}
\begin{figure}[htbp]
            \centering
            \begin{subfigure}[b]{0.48\textwidth}
                \centering
                \includegraphics[width=\textwidth]{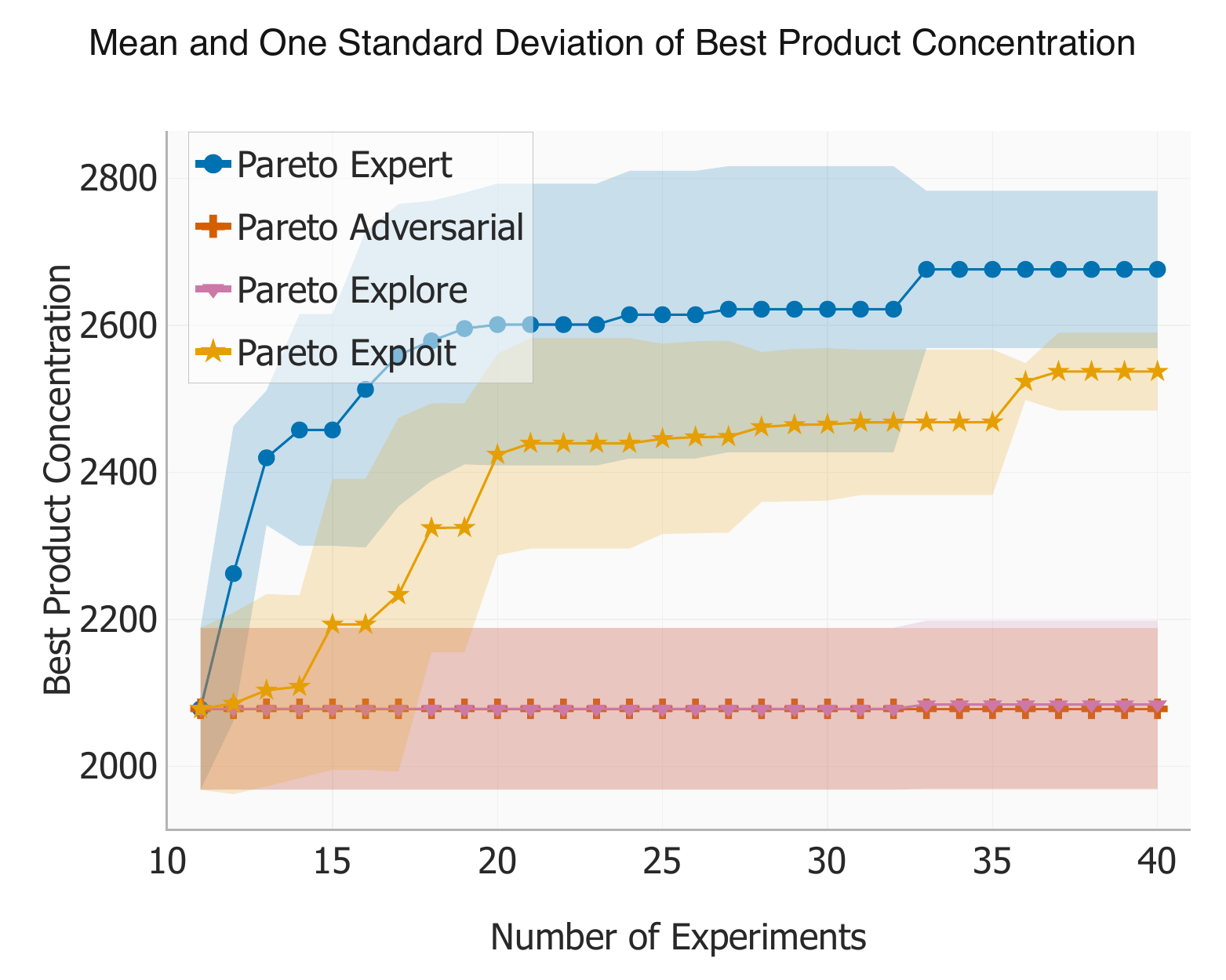}
                \caption{}
                \label{fig:persona_performance}
            \end{subfigure}
            \hfill
            \begin{subfigure}[b]{0.48\textwidth}
                \centering
                \includegraphics[width=\textwidth]{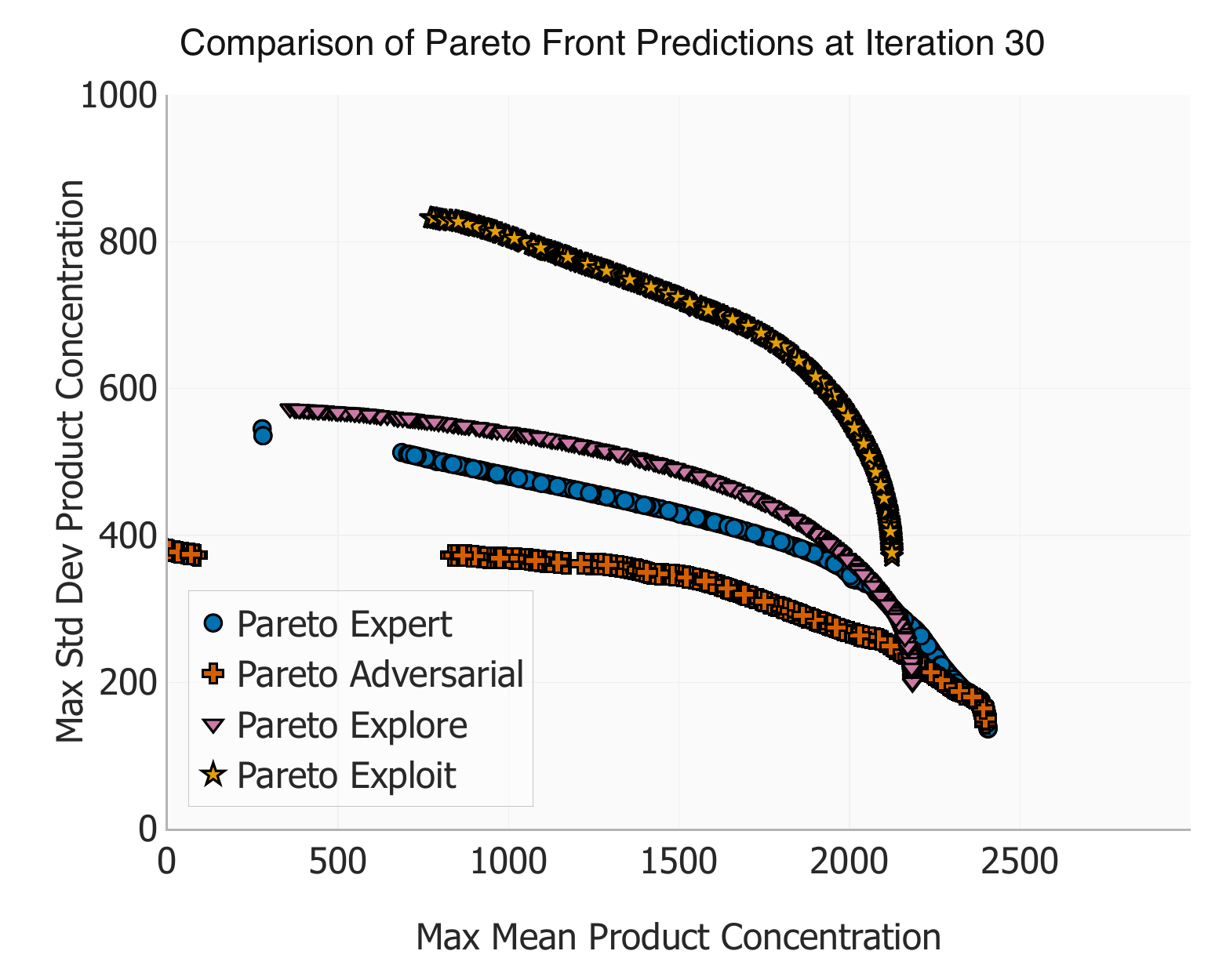}
                \caption{}
                \label{fig:persona_pareto}
            \end{subfigure}
            \caption{Comparison of persona performance and the corresponding Pareto fronts over 5 random initializations and 30 iterations. (a) Best experimental titer achieved by each of the four expert personas (Pareto Explore, Pareto Exploit, Pareto Expert, Pareto Adversarial) over the course of the optimization. (b) Pareto fronts of the four personas after 30 iterations, the adversarial persona identifies high-performing recipes however, they were never experimentally tested.}
            \label{fig:persona}
        \end{figure}

        In order to study the effect of the expert in the proposed methodology, four personas were created:
        \begin{itemize}
            \item \textbf{Pareto Explore}: This persona always takes the most explorative recipe (highest standard deviation) from the Pareto front.
            \item \textbf{Pareto Exploit}: This persona always takes the most exploitative recipe (highest mean prediction) from the Pareto front.
            \item \textbf{Pareto Expert}: This persona always takes the recipe from the Pareto front that has the true best objective value. Note that this is a fictitious scenario in which we assume an expert with perfect knowledge, obtained by using data leakage to generate this persona.
            \item \textbf{Pareto Adversarial}: This persona always takes the recipe from the Pareto front that has the true worst objective value. As with the expert persona, this is a fictitious scenario using data leakage to simulate perfect, but adversarial, knowledge.
        \end{itemize}
        The performance of the personas for 5 random initializations each is shown in Figure~\ref{fig:persona}. As expected, the Pareto Expert performs best, followed by the exploitative persona. Both the explorative and adversarial personas perform poorly. However, Figure~\ref{fig:persona_pareto} shows the Pareto fronts of the different algorithms after 30 iterations, and it can be seen that the adversarial persona has identified high-performing recipes with the same uncertainty as the expert persona, but has not run an experiment at that position, which is why its performance appears worse in Figure~\ref{fig:persona_performance}. This shows that although the adversarial persona chooses the worst possible point, the model is still able to learn from the experiments, and since it always selects from a Pareto front, it is, by definition, always choosing an optimal point given the trade-off between exploration and exploitation.

        A further comparison to \ac{BO} can be found in~\cite{stricker_pareto_2026} and a comparison to the personas in~\ref{appx:bo_compare}.

%% file: 04_conclusion.tex
\section{Conclusion}
    The results highlight several practical advantages of this approach for process development. Expert-defined requirements can restrict the search to feasible and relevant regions, while the Pareto-front representation provides insight into the potential performance within these regions. This also supports informed stopping decisions: if the remaining feasible candidates no longer provide sufficient improvement, the expert can decide that the current region has been characterized with adequate accuracy. Although \ac{GP} models were used in this case study, the framework is not limited to \acp{GP}. Any probabilistic surrogate model that provides reliable estimates of a probability distribution could, in principle, be used within the same \ac{PFGS} workflow. Similarly, while the case study focused on biopharmaceutical process development, the methodology is applicable more broadly to experimental optimization problems where expert knowledge, constraints, and robustness are important.

    Several limitations remain. Introducing a human decision-maker also introduces the possibility of bias or suboptimal guidance, especially if the selected requirements do not reflect the true process-development objective. The quality of the final recommendations therefore depends not only on the surrogate model, but also on the expertise and consistency of the user. In addition, the reliability of the Pareto front depends on the convergence of the underlying optimization algorithm. If the genetic algorithm (\ac{NSGAII}) does not adequately approximate the true non-dominated front, relevant candidate recipes may be missed or incorrectly represented.
    
    Future work should extend the framework to multi-fidelity optimization, where experiments at different scales or levels of accuracy can be balanced against their respective cost and information gain~\cite{lima_multi-fidelity_2025}. This is particularly relevant for scale-up workflows, where small-scale experiments, pilot-scale studies, and full-scale validation runs differ substantially in cost and reliability~\cite{martens_holistic_2025}. Additional case studies in chemistry, chemical engineering, and other experimental sciences should also be investigated to assess the generality of the approach. Finally, future work should compare \ac{HitL} \ac{PFGS} with fully automated \ac{BO} strategies in controlled benchmark studies. Once the surrogate model has reached sufficient accuracy, the objective could also be reformulated for late-stage process optimization and manufacturing, where the goal is no longer to explore uncertain regions but to identify robust, constraint-satisfying operating conditions with low predictive uncertainty.

%% file: 05_acknowledgments.tex
\section*{Author Contributions}
    \textbf{Samuel Stricker:} Conceptualization; methodology; software; writing -- original draft; investigation; formal analysis; visualization.\\
    \textbf{Claus Wirnsperger:} Conceptualization; methodology; writing -- review and editing.\\
    \textbf{Alessandro Butt\'e:} Resources; writing -- review and editing; supervision.\\
    \textbf{Laura Marie Helleckes:} Writing -- review and editing; methodology; conceptualization.\\
    \textbf{Gonzalo Guill\'en Gos\'albez:} Resources; supervision; writing -- review and editing.\\
    \textbf{Antonio del Rio Chanona:} Supervision; methodology; conceptualization; writing -- review and editing.\\
    \textbf{Mehmet Mercang\"oz:} Conceptualization; methodology; supervision; writing -- review and editing; project administration.

\section*{Acknowledgments}
    This research was initiated as part of a Master's thesis at ETH Zurich in collaboration with DataHow AG and was subsequently further developed during an internship at DataHow AG and doctoral studies at Imperial College London. The work was supported by financial contributions from DataHow AG and the Department of Chemical Engineering Scholarship at Imperial College London. Helleckes L. M. is supported by an Eric and Wendy Schmidt AI in Science Postdoctoral Fellowship, a Schmidt Sciences program. Generative AI tools were used to assist with language editing and coding, and all content was reviewed and approved by the authors.

\section*{Funding Information}
    The work was supported by financial contributions from DataHow AG and the Department of Chemical Engineering Scholarship at Imperial College London. Helleckes L. M. is supported by an Eric and Wendy Schmidt AI in Science Postdoctoral Fellowship, a Schmidt Sciences program.

\section*{Conflict of Interest Statement}
    The authors declare no conflict of interest. C.W. and A.B. are employees of DataHow AG, which co-funded this work through a collaboration with Imperial College London.

\section*{Data Availability Statement}
    The code implementing PFGS, including the constrained and robust extensions and the bioprocess simulator used in the case study, will be made available upon publication on the Autonomous Industrial Systems Lab GitHub page (\url{https://github.com/AISL-at-Imperial-College-London}).

\section*{ORCID}
    \textbf{Samuel Stricker:} \href{https://orcid.org/0009-0008-2261-8430}{0009-0008-2261-8430}\\
    \textbf{Claus Wirnsperger:} \href{https://orcid.org/0009-0000-8511-1428}{0009-0000-8511-1428}\\
    \textbf{Alessandro Butt\'e:} \href{https://orcid.org/0000-0003-3506-3792}{0000-0003-3506-3792}\\
    \textbf{Laura Marie Helleckes:} \href{https://orcid.org/0000-0001-7825-7998}{0000-0001-7825-7998}\\
    \textbf{Gonzalo Guill\'en Gos\'albez:} \href{https://orcid.org/0000-0001-6074-8473}{0000-0001-6074-8473}\\
    \textbf{Antonio del Rio Chanona:} \href{https://orcid.org/0000-0003-0274-2852}{0000-0003-0274-2852}\\
    \textbf{Mehmet Mercang\"oz:} \href{https://orcid.org/0000-0002-4449-0414}{0000-0002-4449-0414}

%% file: 06_appendix.tex
\renewcommand{\appendixname}{Supporting Information}

\appendix

\setcounter{figure}{0}
\renewcommand{\thefigure}{SI \arabic{figure}}
\renewcommand{\figurename}{Fig}

\setcounter{table}{0}
\renewcommand{\thetable}{SI \arabic{table}}
\renewcommand{\tablename}{Table}

\setcounter{equation}{0}
\renewcommand{\theequation}{SI \arabic{equation}}

\renewcommand{\thesection}{SI \arabic{section}}
\section*{Supporting Information}
\acresetall % the SI is self-contained: expand acronyms again at first use

    \section{Bayesian Optimization}\label{appx:BO} 
    \subsection{Gaussian Processes}\label{sec:gp}
        \acp{GP} are used for many \ac{BO} workflows because they combine flexible non-parametric regression with explicit uncertainty estimates. A \ac{GP} defines a distribution over possible latent functions rather than a single fitted response surface. It is fully specified by a mean function $m(\mathbf{x})$ and a covariance, or kernel, function $k(\mathbf{x}, \mathbf{x}')$. We denote this as:
        \begin{equation}
            f(\mathbf{x}) \sim \mathcal{GP}\!\left(m(\mathbf{x}), k(\mathbf{x}, \mathbf{x}')\right).
        \end{equation}
        The mean function represents the expected function value, while the kernel encodes assumptions about similarity between input conditions, such as smoothness, stationarity, periodicity, or characteristic length scales~\cite{frazier_tutorial_2018}.
    
        Given observations $\mathcal{D}_{t}$, where measurements are modeled as $y_i = f(\mathbf{x}_i) + \epsilon_i$ with Gaussian noise $\epsilon_i \sim \mathcal{N}(0, \sigma_\epsilon^2)$, Bayesian conditioning gives a posterior predictive distribution at a new candidate point $\mathbf{x}_*$:
        \begin{equation}
            p(f_* \mid \mathbf{x}_*, \mathcal{D}_t) = \mathcal{N}\!\left(\mu_{n_d}(\mathbf{x}_*), \sigma_{n_d}^2(\mathbf{x}_*)\right).
        \end{equation}
        For a zero-mean prior, the posterior mean and variance are
        \begin{align}
            \mu_n(\mathbf{x}_*) &= \mathbf{k}_*^\top (\mathbf{K} + \sigma_\epsilon^2 \mathbf{I})^{-1} \mathbf{y}, \\
            \sigma_n^2(\mathbf{x}_*) &= k(\mathbf{x}_*, \mathbf{x}_*) - \mathbf{k}_*^\top (\mathbf{K} + \sigma_\epsilon^2 \mathbf{I})^{-1} \mathbf{k}_*,
        \end{align}
        where $\mathbf{K}=k(\mathbf{X},\mathbf{X})$ is the covariance matrix of the observed inputs, $\mathbf{k}_*=k(\mathbf{X},\mathbf{x}_*)$, and $\mathbf{y}$ contains the measured responses. The posterior mean acts as the expected best estimate of process performance, while the posterior variance can be used to quantify remaining uncertainty~\cite{rasmussen_gaussian_2005}. This dual output makes \acp{GP} particularly suitable as \ac{BO} surrogates, since acquisition functions can use $\mu_n(\mathbf{x})$ to exploit promising regions and $\sigma_n(\mathbf{x})$ to explore uncertain regions.
    
        In bioprocess development, \acp{GP} are attractive because they are data-efficient, provide calibrated uncertainty, and can be combined with mechanistic knowledge. They have been used, for example, in hybrid dynamic time-series models for fed-batch cultures~\cite{cruz-bournazou_hybrid_2022} and are commonly discussed as surrogate models for data-efficient optimization of media, feeding strategies, and other expensive bioprocess design tasks~\cite{gisperg_bayesian_2025}.
    
    \subsection{Acquisition Functions}\label{sec:aquisition}
        Acquisition functions translate the \ac{GP} posterior into a decision rule for selecting the next experiment. They assign each candidate condition a scalar utility that reflects the expected value of sampling there, thereby balancing exploitation of regions with high predicted performance against exploration of regions with high uncertainty. Improvement-based criteria such as \ac{EI} and \ac{PI} quantify the gain relative to the best response observed so far. For a maximization problem, they are commonly written as
        \begin{align}
            \alpha_{\text{EI}}(\mathbf{x} \mid \mathcal{D}_{t}) 
                &= \mathbb{E}\!\left[\max(f(\mathbf{x}) - f^{*}, 0)\right] 
                = (\mu(\mathbf{x}) - f^{*})\,\Phi(Z) + \sigma(\mathbf{x})\,\phi(Z), \\
            \alpha_{\text{PI}}(\mathbf{x} \mid \mathcal{D}_{t}) 
                &= P(f(\mathbf{x}) > f^{*}) 
                = \Phi(Z),
        \end{align}
        where $f^{*}$ is the current best observation, $Z = \frac{\mu(\mathbf{x}) - f^{*}}{\sigma(\mathbf{x})}$, and $\Phi(\cdot)$ and $\phi(\cdot)$ denote the standard normal Cumulative Distribution Function (\acs{CDF}) and \ac{PDF}\acused{CDF}, respectively~\cite{frazier_tutorial_2018}. \ac{EI} tends to favor points that either have a high posterior mean or a substantial chance of exceeding the incumbent, whereas \ac{PI} focuses only on the probability of improvement and can therefore be more exploitative. Confidence-bound acquisitions provide an alternative view by explicitly adding or subtracting an uncertainty bonus. For maximization, \ac{UCB} uses optimism under uncertainty:
        \begin{align}
            \alpha_{\text{UCB}}(\mathbf{x} \mid \mathcal{D}_{t}) &= \mu_n(\mathbf{x}) + \beta\,\sigma_n(\mathbf{x}), 
            % \alpha_{\text{LCB}}(\mathbf{x} \mid \mathcal{D}_{t}) &= \mu_n(\mathbf{x}) - \beta\,\sigma_n(\mathbf{x}).
            \label{eq:confidence_bounds}
        \end{align}
        The parameter $\beta > 0$ controls the exploration--exploitation trade-off: larger values place more emphasis on uncertain regions, while smaller values focus the search near the current posterior optimum~\cite{srinivas_information-theoretic_2012}.
    
        In practice, the acquisition function is important because it determines how limited experimental capacity is allocated across uncertain and potentially noisy design spaces. In these settings, acquisition functions provide a principled alternative to fixed experimental plans by using each new observation to update the surrogate model and adaptively prioritize the next most informative experiments~\cite{siska_guide_2026}. However, most acquisition functions like \ac{UCB} (Equation~\ref{eq:confidence_bounds}) have hyperparameter that are often difficult to define in practice and have a significant impact to the performance of \ac{BO}. We can see in Section~\ref{sec:pfgs} that \ac{PFGS} removes this parameters and replaces them with easier to estimate, domain specific criteria. Additionally, incorporating expert knowledge into the acquisition function can be challenging, especially when the expert's preferences are difficult to quantify or may change over time. 

    \subsection{Batch Bayesian Optimization}
        Parallel evaluation of multiple experimental conditions is common in modern bioprocess development, for example when microtiter plates, parallel bioreactors, or automated liquid-handling platforms are available. Batch \ac{BO} extends the sequential \ac{BO} loop by proposing a set of $n_q$ candidate points at each iteration rather than a single next experiment. The objective is not only to select individually promising conditions, but also to choose a batch that is informative as a whole. This requires balancing exploitation, exploration, and diversity within the batch, since selecting several nearby points can waste experimental capacity by providing redundant information.

        Batch acquisition functions therefore evaluate the joint utility of a candidate set $\mathbf{X}_{n_q}=\{\mathbf{x}_{1},\dots,\mathbf{x}_{n_q}\}$. Common examples include Monte Carlo variants of expected improvement and confidence-bound criteria, such as $q$-\ac{EI} and $q$-\ac{UCB}~\cite{balandat_botorch_2020}. $q$-\ac{EI} generalizes \ac{EI} by considering the best improvement achieved by any point in the proposed batch. Unlike standard \ac{EI}, $q$-\ac{EI} has no simple closed form for $n_q>1$ because it depends on the maximum of correlated posterior predictions. It is therefore typically approximated by drawing $S$ joint posterior samples and averaging the improvement across samples:
        \begin{align}
            \alpha_{\text{q-EI}}(\mathbf{X}_{n_q} \mid \mathcal{D}_{t})  
            &= \mathbb{E}\!\left[\max_{j=1,\dots,n_q}\max\!\left(f(\mathbf{x}_{j}) - f^{*},\, 0\right)\right]\\
            &\approx \frac{1}{n_s}\sum_{s=1}^{n_s}\max_{j}\max\!\left(f^{(s)}(\mathbf{x}_{j}) - f^{*},\, 0\right).
        \end{align}
        Alternative batch strategies include local penalization, which sequentially constructs a batch by down-weighting acquisition values near already selected points~\cite{gonzalez_batch_2016}, and Thompson sampling, where multiple draws from the posterior naturally induce diversity among proposed experiments~\cite{kandasamy_parallelised_2018}. These approaches can be computationally lighter than fully joint Monte Carlo acquisition optimization since they construct the batch sequentially instead of maximizing its joint value, at the cost of no guarantee that the chosen set is jointly optimal. Local penalization also underpins asynchronous batch designs, where new experiments must be proposed before all running experiments have returned. Folch et al.~\cite{folch_combining_2023} combine this idea with multi-fidelity modelling in an acquisition-function-agnostic framework, showing that asynchronous multi-fidelity batches outperform both single-fidelity batch and multi-fidelity sequential baselines on a battery-electrode design task where coin-cell experiments approximate pouch-cell performance.

    \subsection{Constrained and Safe Bayesian Optimization} \label{sec:constrained_bo}
        Many optimization problems are defined not only by a performance objective, but also by feasibility requirements. Two related paradigms address this setting, differing mainly in \emph{when} feasibility or safety must hold. Constrained \ac{BO} aims to identify a final solution that satisfies the constraints, whereas safe \ac{BO} requires every evaluated query during the optimization campaign to satisfy the safety constraints with high probability. 

        Constrained \ac{BO} considers an expensive black-box objective subject to one or more unknown inequality constraints,
        \begin{equation}
            c_k(\mathbf{x}) \leq 0,\qquad k=1,\dots,n_{c},
        \end{equation}
        where the feasible region is learned from data and may be nonlinear or disconnected. A common approach is to fit a separate \ac{GP} surrogate to each constraint in addition to the objective surrogate. The acquisition function then combines objective improvement with the posterior probability that all constraints are satisfied~\cite{gelbart_bayesian_2014}. Modeling the objective and each constraint with independent \ac{GP} surrogates, the feasibility probability factorizes across constraints, and constrained expected improvement becomes
        \begin{equation}
            \alpha_{\text{cEI}}(\mathbf{x} \mid \mathcal{D}_{t})
            = \alpha_{\text{EI}}(\mathbf{x} \mid \mathcal{D}_{t})
            \prod_{k=1}^{n_{c}} P\!\left(c_k(\mathbf{x}) \leq 0\right),
        \end{equation}
        where, under a Gaussian posterior for constraint $k$, the feasibility probability is
        \begin{equation}
            P\!\left(c_k(\mathbf{x}) \leq 0\right)
            = \Phi\!\left(\frac{-\mu_{c_k}(\mathbf{x})}{\sigma_{c_k}(\mathbf{x})}\right).
        \end{equation}
        This formulation favors candidates that are both likely to improve the objective and likely to satisfy the constraints~\cite{gelbart_bayesian_2014}. If no feasible point has yet been observed, the improvement term may collapse. In this early phase, constrained \ac{BO} is often driven by feasibility search, selecting points that maximize $\prod_{k=1}^{n_k}P(c_k(\mathbf{x})\leq 0)$ until at least one feasible observation is identified.

        In safety-critical systems, however, allowing infeasible evaluations during the search may be unacceptable or irreversible. Safe \ac{BO} addresses this stricter setting by requiring all queried points to satisfy a safety constraint with high probability throughout the optimization process, rather than only at convergence~\cite{fiedler_safety_2024}.
        
        In bioprocess engineering, constrained \ac{BO} is particularly relevant because \ac{CQA} violations are typically feasibility concerns that need to be addressed. This makes constrained \ac{BO} closely aligned with Quality by Design and systematic process development. It allows the optimizer to search for high titre, purity, or yield while explicitly accounting for \ac{CQA} limits, viable operating windows, and regulatory or manufacturing constraints.

    \subsection{Adversarially Robust Bayesian Optimization} \label{sec:rob_bo}
        Classical \ac{BO} seeks an input with high predicted objective value, but the resulting optimum may be sensitive to small deviations between the proposed and realized operating conditions. This is problematic in experimental process development, where implementation errors, raw-material variability, or scale-dependent effects can perturb the selected recipe. Robust \ac{BO} therefore modifies the optimization goal from finding a single high-performing point to finding a point whose performance remains high under a specified perturbation set.

        Bogunovi\'c et al.~\cite{bogunovic_adversarially_2018} formulate this idea as adversarially robust \ac{BO}. Let $d(\mathbf{x},\mathbf{x}')$ denote a distance measure on the domain $\mathcal{X}$, and let $\epsilon$ define the size of the admissible perturbation. For a nominal point $\mathbf{x}$, the local perturbation set is
        \begin{equation}
            \Delta_{\epsilon}(\mathbf{x})
            = \{\boldsymbol{\delta}:\mathbf{x}+\boldsymbol{\delta}\in\mathcal{X},\, d(\mathbf{x},\mathbf{x}+\boldsymbol{\delta})\leq\epsilon\}.
        \end{equation}
        The robust optimum is then defined by a max--min objective,
        \begin{equation}
            \mathbf{x}^{*}_{\epsilon}
            \in \arg\max_{\mathbf{x}\in\mathcal{X}}
            \min_{\boldsymbol{\delta}\in\Delta_{\epsilon}(\mathbf{x})}
            f(\mathbf{x}+\boldsymbol{\delta}),
        \end{equation}
        which favors broad, stable regions of high performance rather than narrow optima that may fail after small perturbations. To solve this problem with \ac{GP} uncertainty, Bogunovi\'c et al. \cite{bogunovic_adversarially_2018} introduce the \textsc{StableOpt} algorithm. Given the \ac{GP} posterior mean $\mu_{{n_d}}(\mathbf{x})$ and standard deviation $\sigma_{n_d}(\mathbf{x})$, the upper and lower confidence bounds are given in Equation~\ref{eq:confidence_bounds} and the algorithm first selects the nominal point with the best robust upper confidence bound,
        \begin{equation}
            \tilde{\mathbf{x}}_{n_d+1}
            \in \arg\max_{\mathbf{x}\in\mathcal{X}}
            \min_{\boldsymbol{\delta}\in\Delta_{\epsilon}(\mathbf{x})}
            \mathrm{ucb}(\mathbf{x}+\boldsymbol{\delta}),
        \end{equation}
        and then samples the most pessimistic perturbation according to the lower confidence bound,
        \begin{equation}
            \boldsymbol{\delta}_{n_d+1}
            \in \arg\min_{\boldsymbol{\delta}\in\Delta_{\epsilon}(\tilde{\mathbf{x}}_n)}
            \mathrm{lcb}(\tilde{\mathbf{x}}_n+\boldsymbol{\delta}).
        \end{equation}
        The experiment is performed at $\tilde{\mathbf{x}}_n+\boldsymbol{\delta}_{n_d}$, and the \ac{GP} posterior is updated with the resulting observation. This combines optimism when choosing a promising robust region with pessimism when identifying the most vulnerable point inside that region.

        In the context of bioprocess development, this formulation is directly relevant because a recommended \ac{CPP} set point is rarely implemented exactly. Robust \ac{BO} provides a principled way to prefer recipes that retain acceptable performance under local \ac{CPP} deviations. This robustness can not only be applied to the objectives but also to the constraints, similar to the work of Ahmed et al.~\cite{ahmed_arrtoc_2025}.

\newpage

    \subsection{\acs{PFGS} vs.\ \acs{BO}} \label{appx:bo_compare}
        \begin{figure}[h!]
           \centering
           \begin{subfigure}{0.48\textwidth}
              \centering
              \includegraphics[width=\textwidth,trim=0 0cm 0 0,clip]{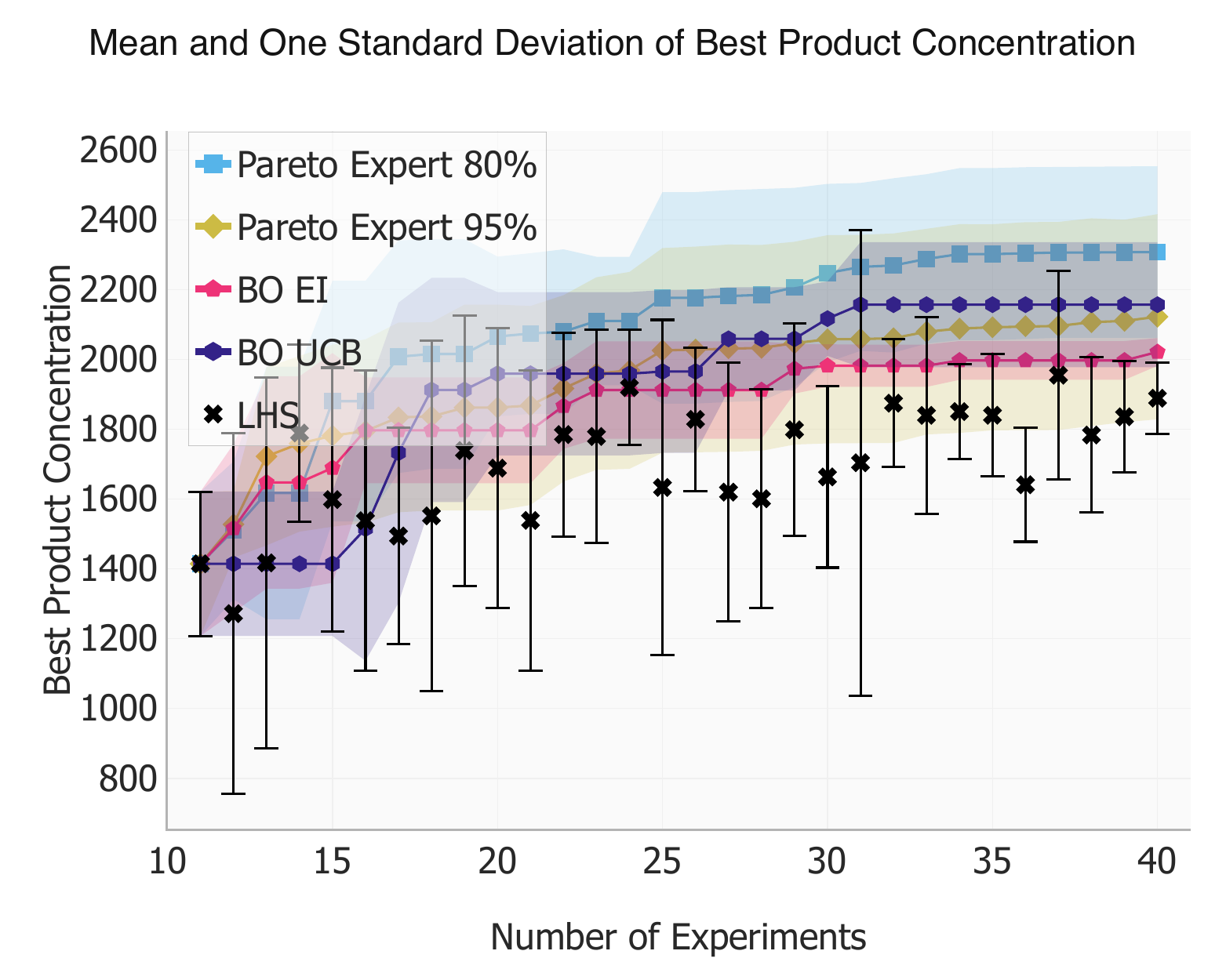}
           \end{subfigure}    
           \begin{subfigure}{0.48\textwidth}
              \centering
              \includegraphics[width=\textwidth,trim=0 0cm 0 0,clip]{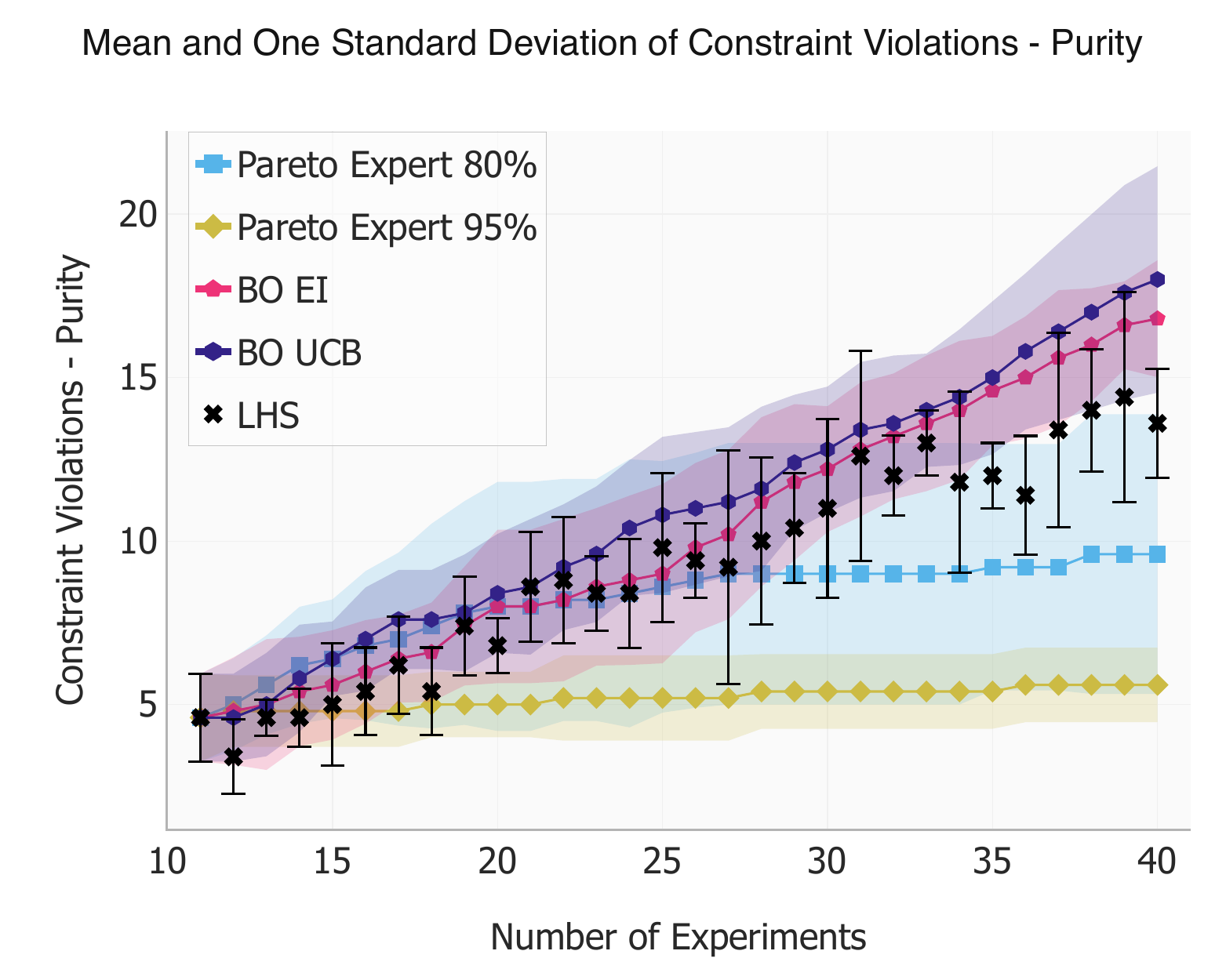}
           \end{subfigure}
            \caption{Performance of the \ac{PFGS} personas compared to \ac{BO} using both the expected improvement and upper confidence bound acquisition functions. The percentages denote the minimum probability of constraint satisfaction required for a point to be selectable by the Pareto Expert. (left) Performance in the objective (titer); as expected, the Expert persona performs comparably to \ac{BO}. (right) Constraint violations of the different optimization strategies; restricting the Pareto front reduces constraint violations, with the degree of reduction depending on how strictly the front is restricted.}
           \label{fig:pfgsbo_constrained}
        \end{figure}
    
    \newpage
    \section{Dashboard Visualization}
    \label{app:dashboard}
        \begin{figure}[h!]
           \centering
           \begin{subfigure}{\textwidth}
              \centering
              \includegraphics[width=0.99\textwidth,trim=0 5cm 0 0,clip]{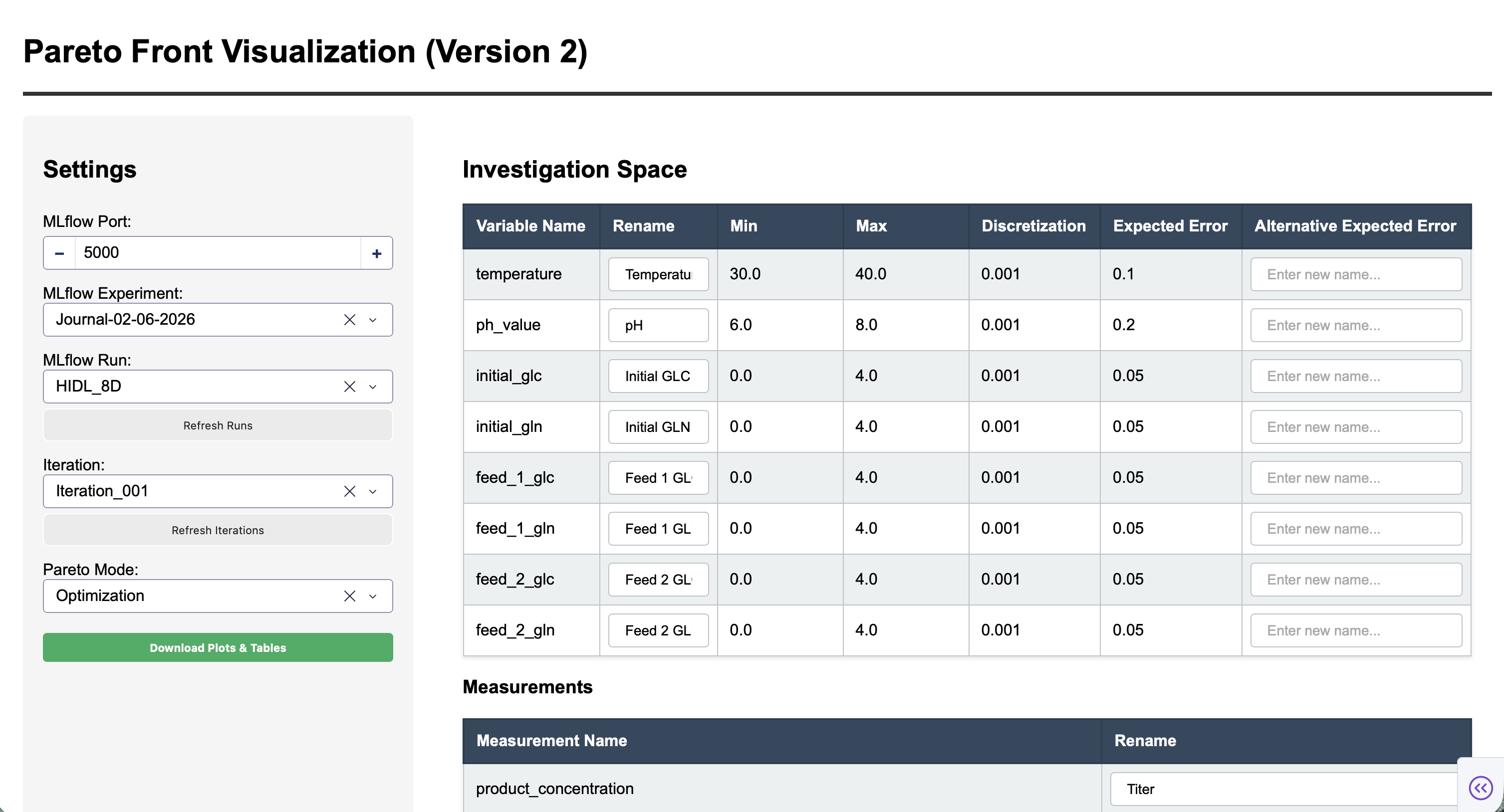}
           \end{subfigure}    
           \begin{subfigure}{\textwidth}
              \centering
              \includegraphics[width=0.99\textwidth,trim=0 7cm 0 0,clip]{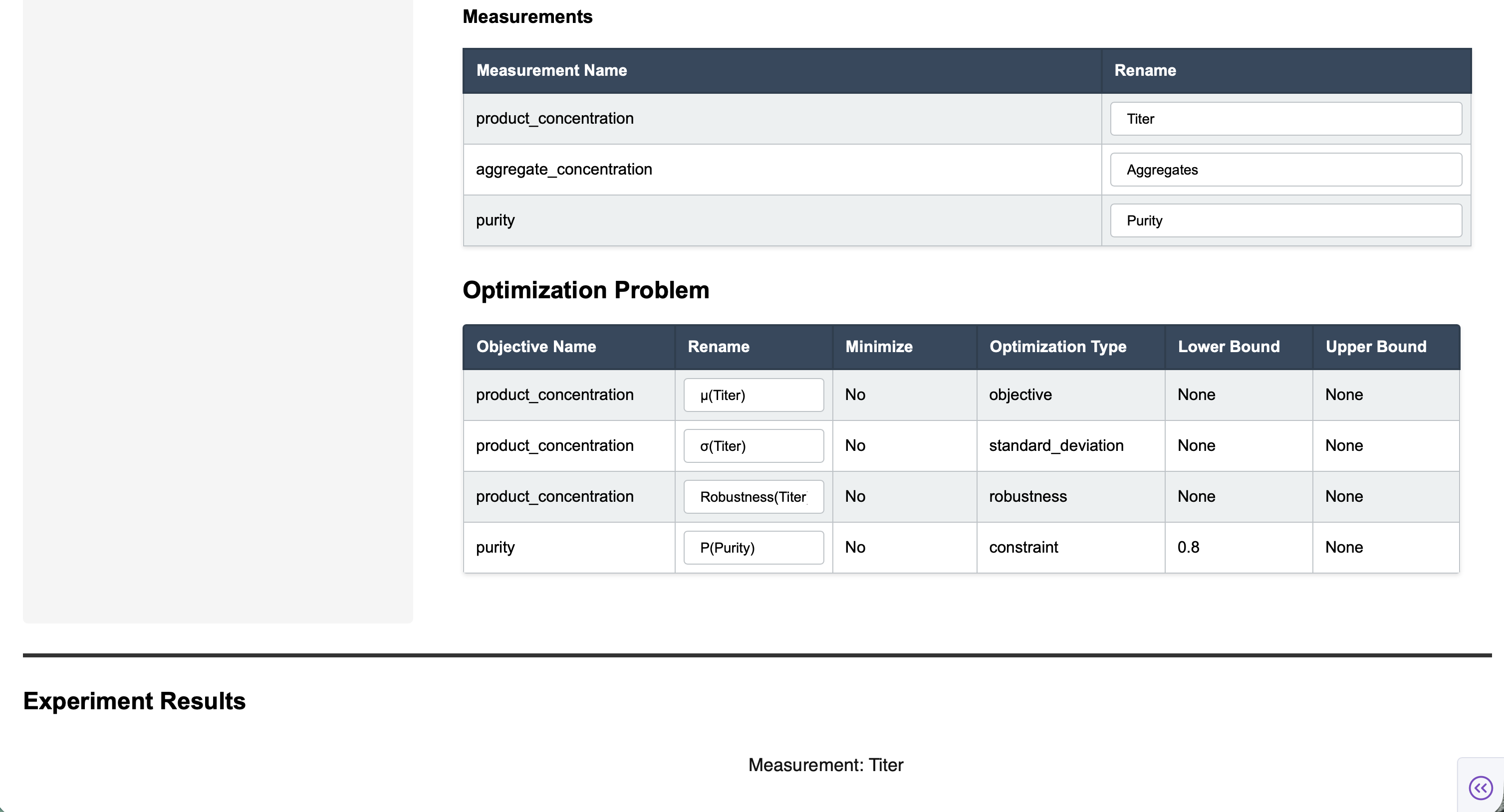}
           \end{subfigure}
            \caption{The setting section loads the relevant run from mlflow and shows the used investigation space and optimization problem.}
           \label{fig:settings}
        \end{figure}
        
         \begin{figure}[h!]
           \centering
           \begin{subfigure}{\textwidth}
              \centering
              \includegraphics[width=0.99\textwidth,trim=0 5cm 0 0,clip]{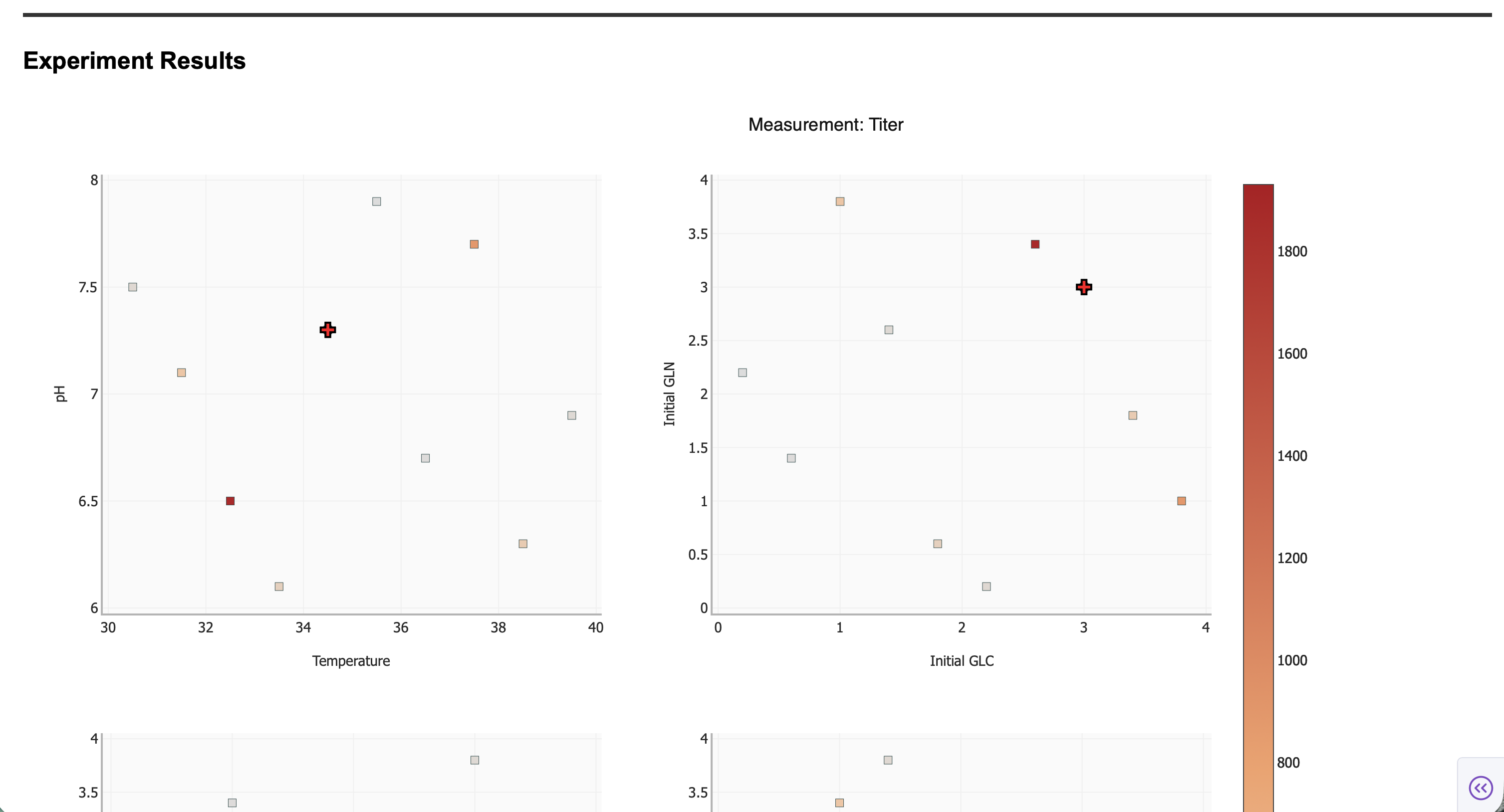}
           \end{subfigure}    
           \begin{subfigure}{\textwidth}
              \centering
              \includegraphics[width=0.99\textwidth,trim=0 3cm 0 0,clip]{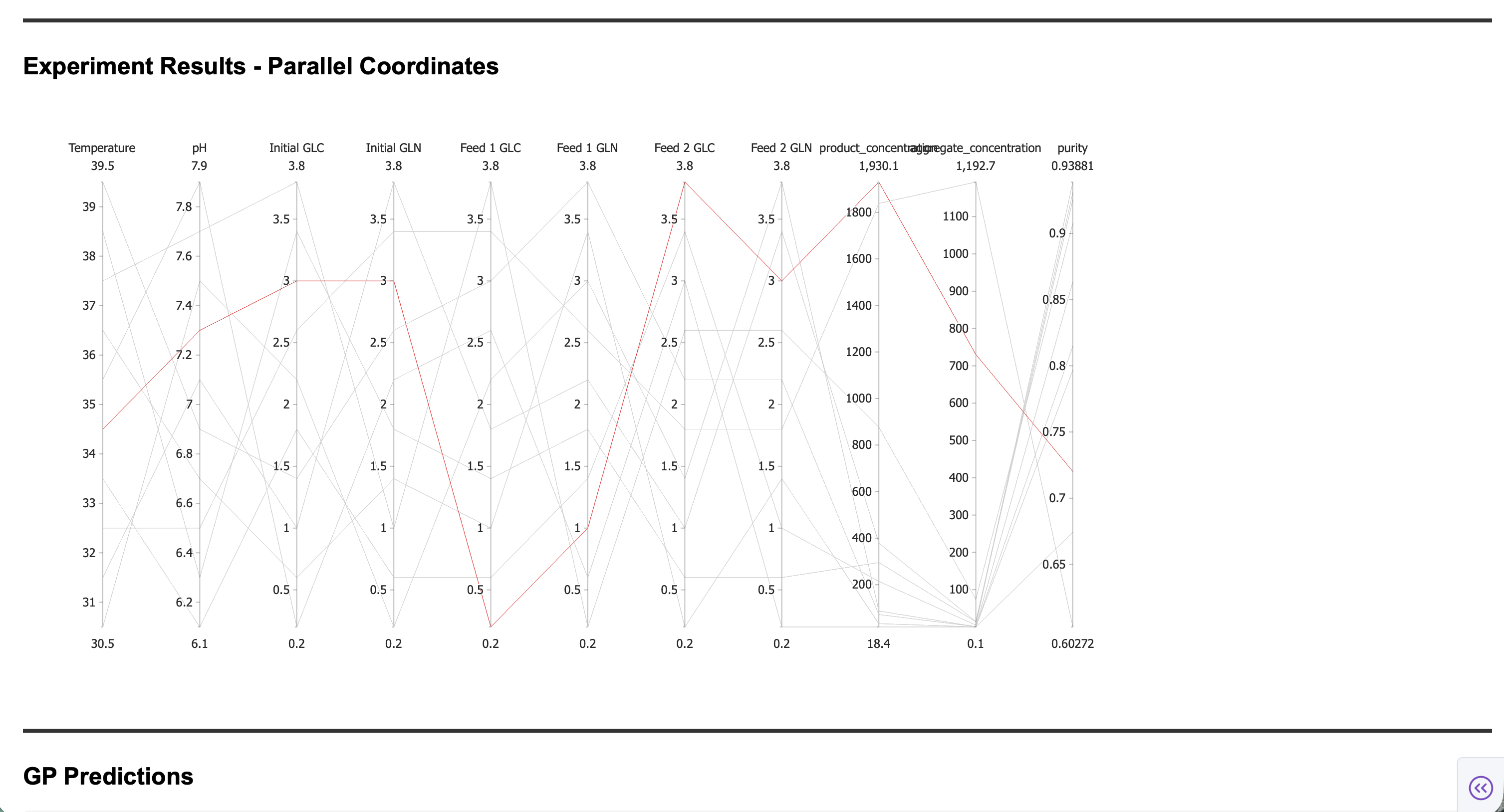}
           \end{subfigure}
            \caption{The Experiment section show the bi-variate investigation space responses of the experiments and the parallel coordinate plot for the performed experiments.}
           \label{fig:ex}
        \end{figure}

        \begin{figure}[h!]
           \centering
           \begin{subfigure}{\textwidth}
              \centering
              \includegraphics[width=0.99\textwidth,trim=0 2cm 0 0,clip]{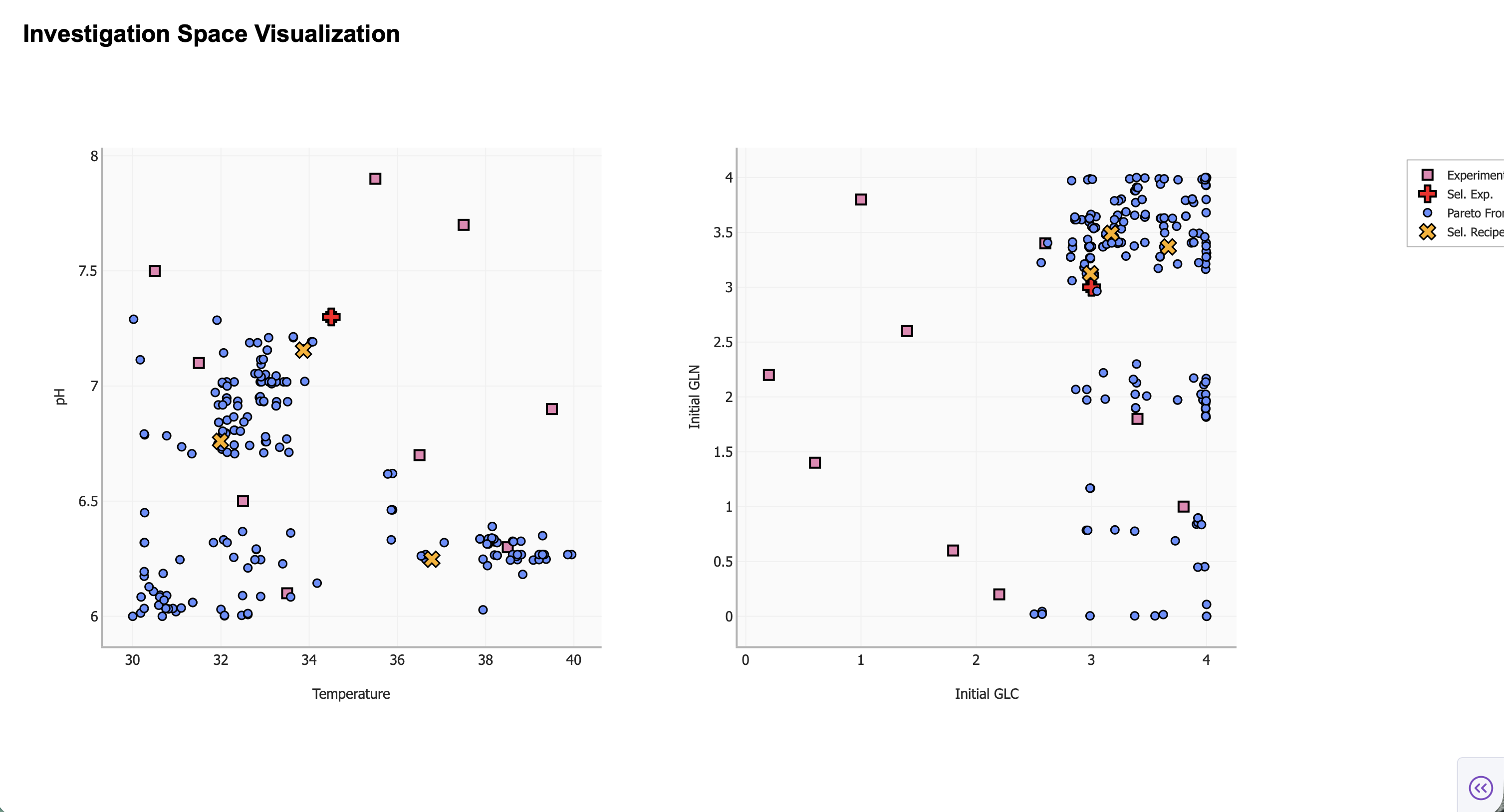}
           \end{subfigure}    
            \caption{The Investigation Space responses of the experiments and the Pareto front.}
           \label{fig:is}
        \end{figure}

        \begin{figure}[h!]
           \centering
           \begin{subfigure}{\textwidth}
              \centering
              \includegraphics[width=0.99\textwidth,trim=0 15cm 0 0,clip]{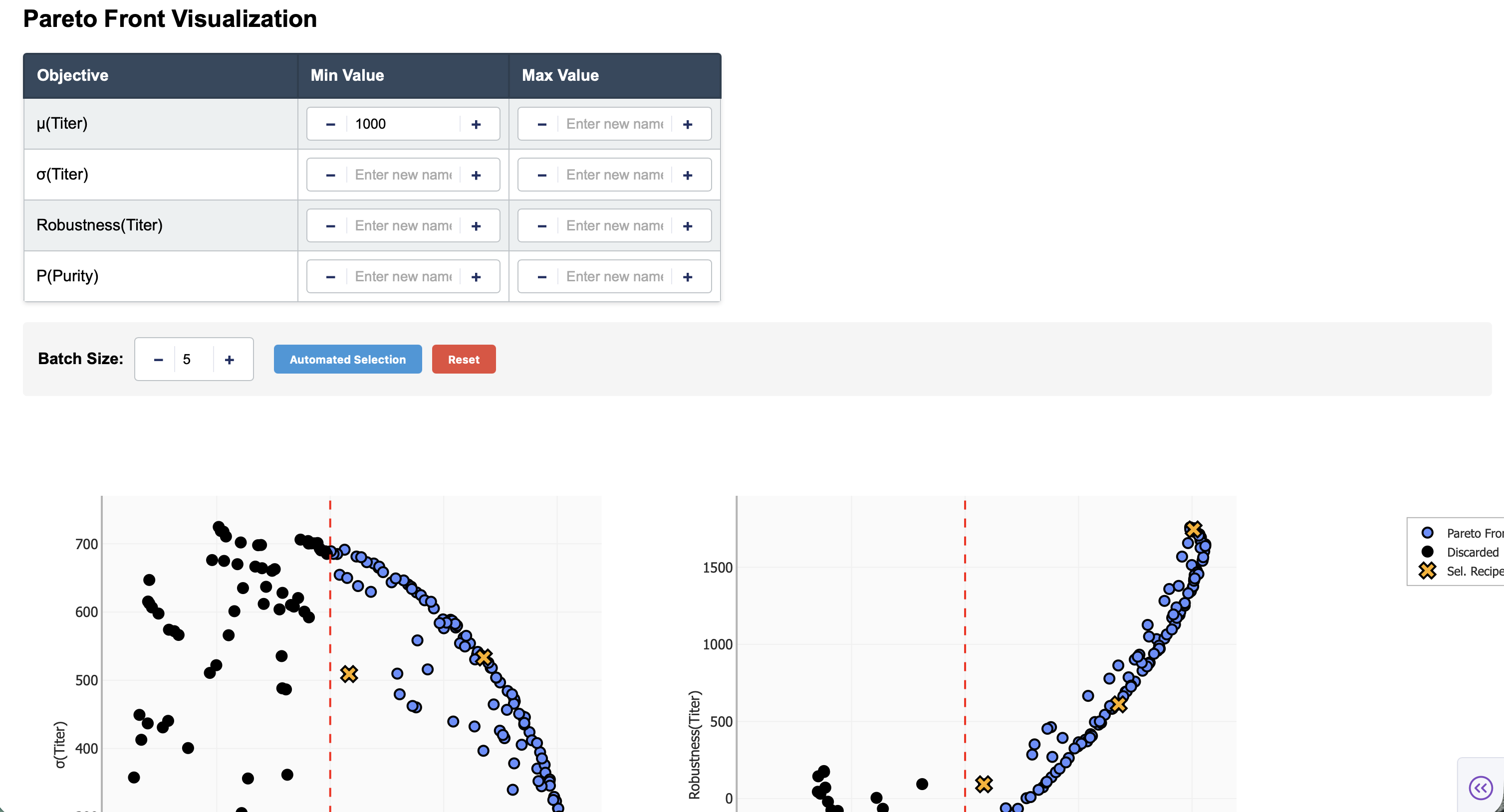}
           \end{subfigure}    
           \begin{subfigure}{\textwidth}
              \centering
              \includegraphics[width=0.99\textwidth,trim=0 5cm 0 1cm,clip]{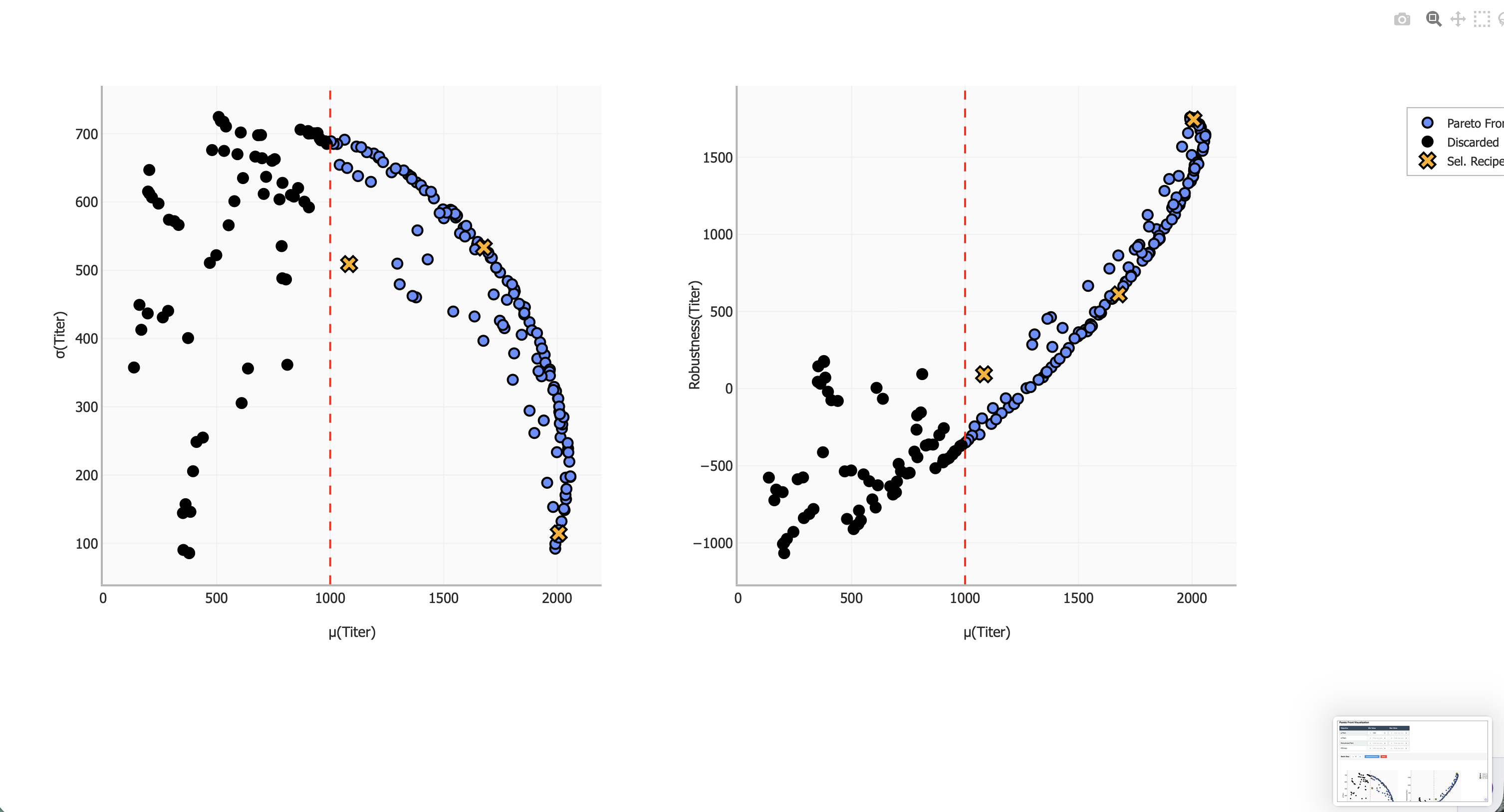}
           \end{subfigure}
           \begin{subfigure}{\textwidth}
              \centering
              \includegraphics[width=0.99\textwidth,trim=0 0cm 0 0,clip]{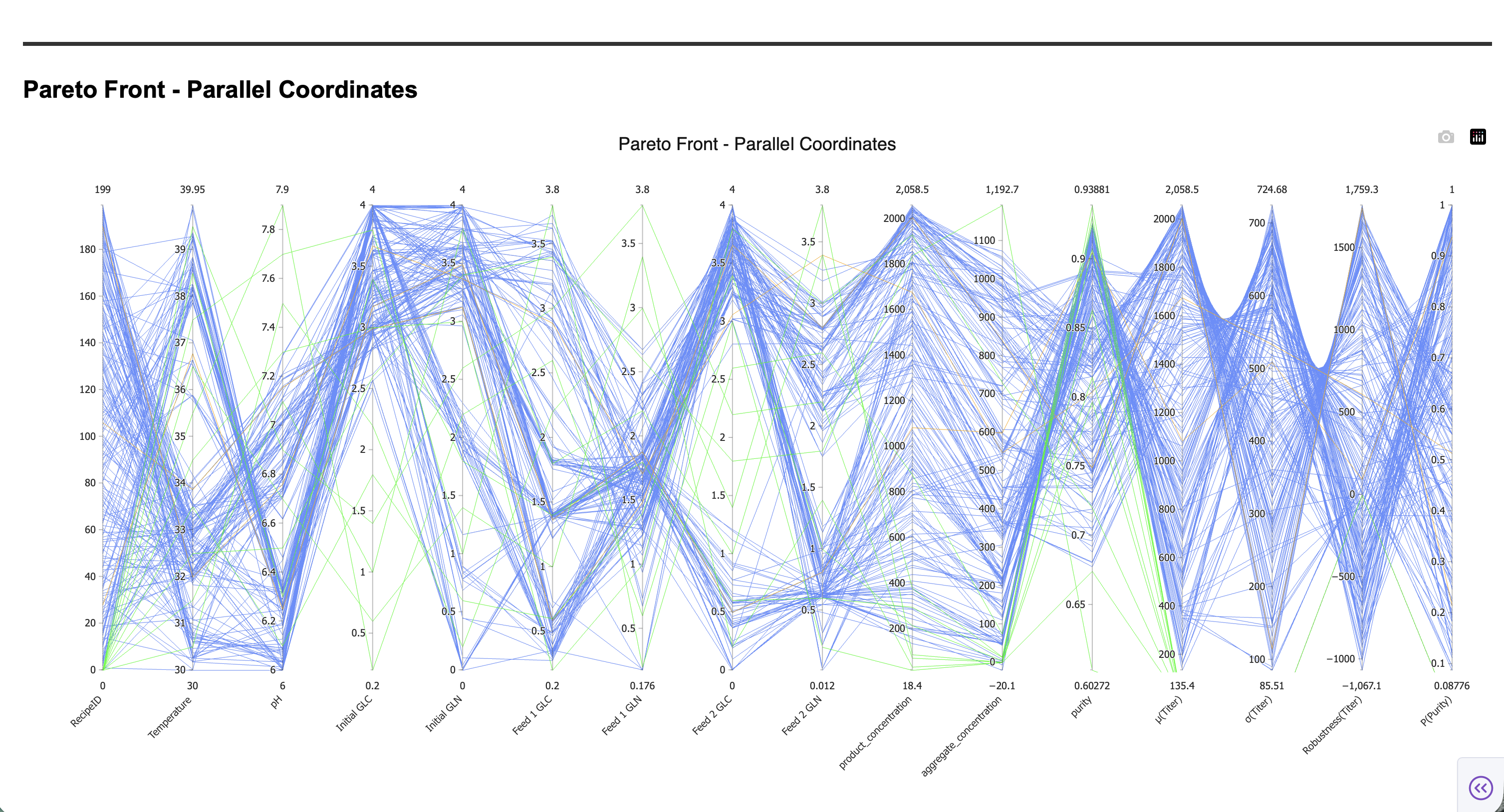}
           \end{subfigure}
            \caption{The Pareto Front section show the bi-variate Pareto front and have the interactive table the restrict the Pareto fronts according to expert requirements.}
           \label{fig:pareto_dash}
        \end{figure}

        \begin{figure}[h!]
           \centering
           \begin{subfigure}{\textwidth}
              \centering
              \includegraphics[width=0.99\textwidth,trim=0 0cm 0 0,clip]{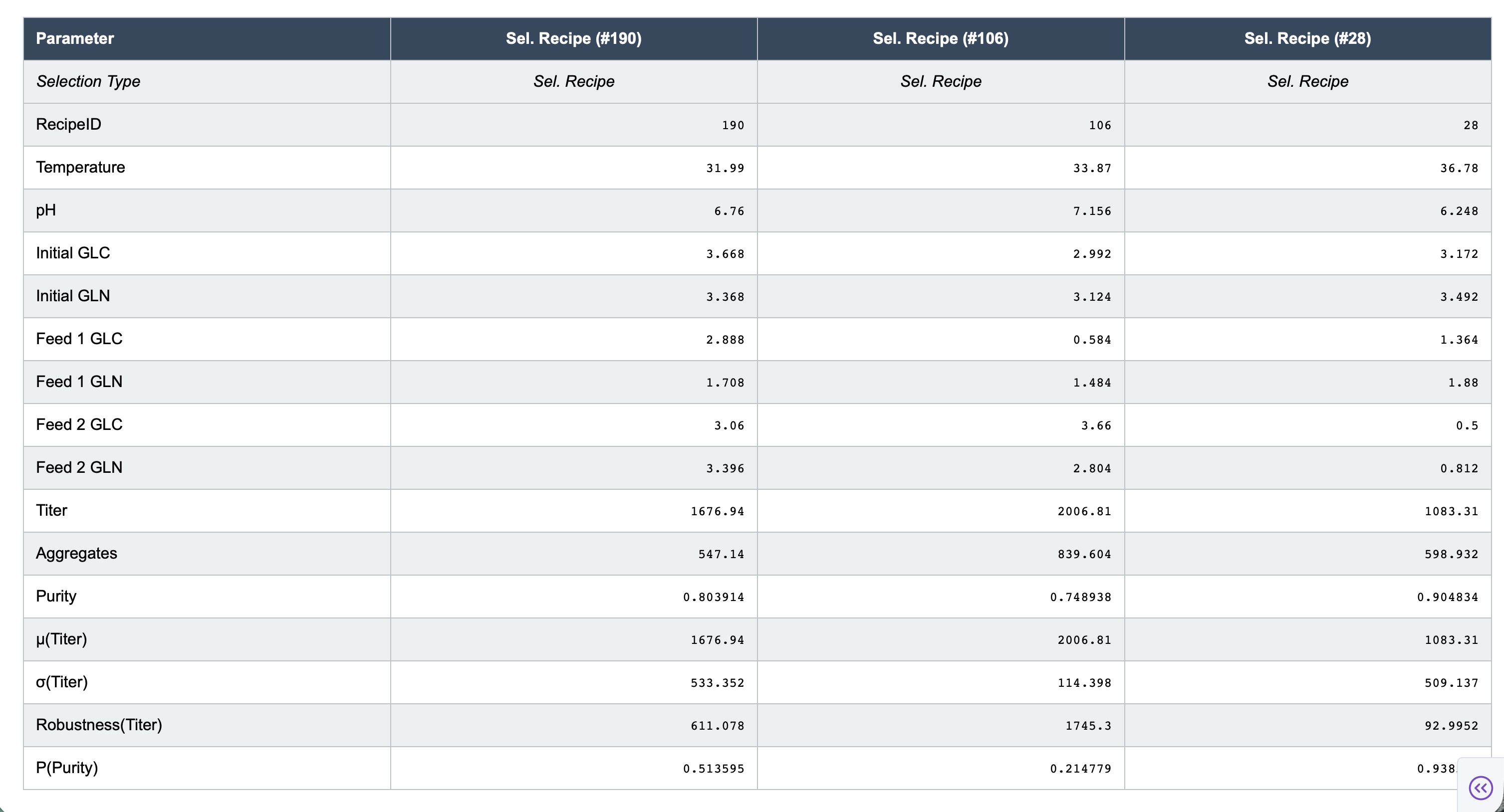}
           \end{subfigure}    
            \caption{The Selected Recipes are shown in a table.}
           \label{fig:select}
        \end{figure}
        
        \FloatBarrier
        
    \section{Bioprocess Simulator}\label{app:bio-simulator}
        The bioprocess simulator models a \ac{CHO} fed-batch culture based on \cite{craven_process_2013, martens_holistic_2025}, extended to include product aggregation and purity. The model describes the dynamics of eight state variables: viable cell density $x_{viable}$, total cell density $x_{total}$, dead cell density $x_{dead}$, product concentration $c_{product}$, aggregate concentration $c_{aggregates}$, glucose $c_{glc}$, glutamine $c_{gln}$, lactate $c_{lac}$, and ammonia $c_{ammonia}$.
        
        \subsection*{Specific Growth Rate}
        
        The specific growth rate is modeled as a product of Monod kinetics and environmental penalty factors:
        
        \begin{equation}
            \mu = \mu_{max} \cdot \frac{c_{glc}}{k_{sat,glc} + c_{glc}} \cdot \frac{c_{gln}}{k_{sat,gln} + c_{gln}} \cdot \frac{k_{sat,lac}}{k_{sat,lac} + c_{lac}} \cdot \frac{k_{sat,ammonia}}{k_{sat,ammonia} + c_{ammonia}} \cdot f_{temp} \cdot f_{pH}
        \end{equation}
        
        where $f_{temp}$ and $f_{pH}$ are asymmetric Gaussian penalty factors for temperature and pH, respectively:
        
        \begin{equation}
            f_{temp} = r_{temp} \cdot \begin{cases}
                \exp\!\left(-0.8 \cdot \left(\dfrac{T - T_{opt}}{3.0}\right)^{\!2}\right) & T < T_{opt} \\[6pt]
                \exp\!\left(-1.0 \cdot \left(\dfrac{T - T_{opt}}{8.0}\right)^{\!2}\right) & T \geq T_{opt}
            \end{cases}
        \end{equation}
        
        \begin{equation}
            f_{pH} = r_{pH} \cdot \begin{cases}
                \exp\!\left(-0.8 \cdot \left(\dfrac{pH - pH_{opt}}{1.5}\right)^{\!2}\right) & pH < pH_{opt} \\[6pt]
                \exp\!\left(-1.5 \cdot \left(\dfrac{pH - pH_{opt}}{0.5}\right)^{\!2}\right) & pH \geq pH_{opt}
            \end{cases}
        \end{equation}
        
        Both factors are clipped to a minimum of 0.2. The effective growth rate incorporates a reactor-specific inhibition term $g_{inhibition}$ and the death rate $k_{death}$:
        
        \begin{equation}
            k_{death} = k_{death,max} \cdot \frac{k_{death,\mu}}{\mu + k_{death,\mu}} \cdot s_{scale}
        \end{equation}
        
        \subsection*{Cell Population Dynamics}
        
        \begin{align}
            \frac{dx_{total}}{dt}  &= \mu \cdot g_{inhibition} \cdot x_{viable} - k_{lysis} \cdot x_{dead} \\
            \frac{dx_{viable}}{dt} &= \left(\mu \cdot g_{inhibition} - k_{death}\right) x_{viable} \\
            \frac{dx_{dead}}{dt}   &= k_{death} \cdot x_{viable} - k_{lysis} \cdot x_{dead}
        \end{align}
        
        \subsection*{Metabolite and Product Dynamics}
        
        \begin{align}
            \frac{dc_{product}}{dt}    &= x_{total} \cdot y_{product,total} + x_{viable} \cdot y_{product,viable} \cdot \mu - 2\,k_{agg}\,c_{product}^2 \\
            \frac{dc_{aggregates}}{dt} &= k_{agg}\,c_{product}^2 \\
            \frac{dc_{glc}}{dt}        &= x_{viable} \left(-\frac{\mu}{y_{cells,glc}} - m_{glc}\right) \\
            \frac{dc_{gln}}{dt}        &= x_{viable} \left(-\frac{\mu}{y_{cells,gln}} - m_{gln}\right) - k_{deg,gln}\,c_{gln} \\
            \frac{dc_{lac}}{dt}        &= -y_{lactate}\,\frac{dc_{glc}}{dt} \\
            \frac{dc_{ammonia}}{dt}    &= -y_{ammonia} \cdot x_{viable} \left(-\frac{\mu}{y_{cells,gln}} - m_{gln}\right) + k_{deg,gln}\,c_{gln}
        \end{align}
        
        \subsection*{Product Purity}
        
        Product purity is defined as the fraction of active product relative to all protein-like and metabolic species:
        
        \begin{equation}
            \text{purity} = \frac{c_{product}}{c_{product} + c_{aggregates} + c_{HCP} + c_{glc} + c_{gln} + c_{lac} + c_{ammonia}}
        \end{equation}
        
        where host cell protein is estimated as $c_{HCP} = x_{dead} \cdot k_{lysis} \cdot hcp_{per\,cell}$.
        
        \subsection*{Noise Model}
        
        Two noise sources are included to reproduce experimental variability. Process noise is modeled as proportional Brownian motion:
        
        \begin{equation}
            dX_t = X_t \cdot \sigma_{process}\sqrt{dt}\,dW_t
        \end{equation}
        
        Measurement noise is modeled as independent proportional Gaussian white noise:
        
        \begin{equation}
            X_{measured} = X_{true}\left(1 + \sigma_{measurement}\,\mathcal{N}(0,1)\right)
        \end{equation}
        
        where $\sigma_{process}$ and $\sigma_{measurement}$ are dimensionless relative noise intensities.

        % The simulator can be downloaded on the groups Github.

    \section{Hyperparameter for Results}
    \label{appx:hyperparam}
        The hyperparameter for \ac{PFGS} used to generate the results in Section~\ref{sec:results}, are listed in Table~\ref{tab:experimental_setup}
        \begin{table}[h]
            \centering
            \caption{Experimental setup for Pareto front optimization.}
            \label{tab:experimental_setup}
            \begin{tabular}{lc}
                \toprule
                Parameter & Value \\
                \midrule
                \multicolumn{2}{l}{\textit{General}} \\
                Initial \ac{LHS} samples & 10 \\
                Batch size & 5 \\
                Robustness confidence & 0.95 \\
                \midrule
                \multicolumn{2}{l}{\textit{\ac{NSGAII}}} \\
                Population size per dimension & 25 \\
                Maximum number of generations & 1000 \\
                \midrule
                \multicolumn{2}{l}{\textit{Bioprocess simulator -- initial conditions}} \\
                $c_{\text{product}}$ & 0.0 \\
                $X_{\text{total}}$ & 800000.0 \\
                $X_{\text{viable}}$ & 800000.0 \\
                \midrule
                \multicolumn{2}{l}{\textit{Bioprocess simulator -- cell type parameters}} \\
                $\mu_{\max}$ & 0.0451639932 \\
                $k_{\text{lysis}}$ & 0.0446990515 \\
                $k_{\text{deg,gln}}$ & 0.0013441722 \\
                $k_{\text{death,max}}$ & 0.0101556537 \\
                $k_{\text{death},\mu}$ & 0.0145125692 \\
                $k_{\text{sat,lac}}$ & 141.4574131165 \\
                $k_{\text{sat,ammonia}}$ & 37.778062819 \\
                $k_{\text{sat,glc}}$ & 1.1634941712 \\
                $k_{\text{sat,gln}}$ & 0.2123961928 \\
                $Y_{\text{ammonia}}$ & $9.98 \times 10^{-8}$ \\
                $Y_{\text{lactate}}$ & 0.8061528449 \\
                $Y_{\text{cells,glc}}$ & $1.5330 \times 10^{8}$ \\
                $Y_{\text{cells,gln}}$ & $8.3637 \times 10^{9}$ \\
                $m_{\text{glc}}$ & $4 \times 10^{-10}$ \\
                $m_{\text{gln}}$ & 0.0 \\
                $\text{pH}_{\text{opt}}$ & 6.7 \\
                pH robustness factor & 1.1 \\
                $E_{\text{a}}$ & 33.4 \\
                Temperature robustness factor & 1.1 \\
                $Y_{\text{product,total}}$ & $9.98 \times 10^{-8}$ \\
                $Y_{\text{product,viable}}$ & $4.9308 \times 10^{-6}$ \\
                HCP per cell & $2 \times 10^{-6}$ \\
                $k_{\text{agg,product}}$ & $3 \times 10^{-6}$ \\
                \bottomrule
            \end{tabular}
        \end{table}